\documentclass[lettersize,journal]{IEEEtran}
\usepackage{amsmath,amsfonts}
\usepackage{amsthm}
\usepackage{algorithmic}
\usepackage[ruled,linesnumbered]{algorithm2e}
\usepackage{array}
\usepackage[caption=false,font=footnotesize,labelfont=rm,textfont=rm]{subfig}
\usepackage{textcomp}
\usepackage{stfloats}
\usepackage{url}
\usepackage{verbatim}
\usepackage{graphicx}
\usepackage{cite}
\usepackage{color, soul} 
\usepackage{booktabs}
\usepackage[top=0.7in,bottom=0.5in,left=0.4in,textwidth=7.7in]{geometry}
\usepackage{multirow}
\usepackage[pagebackref=false,breaklinks=false,linkcolor=red,anchorcolor=black, citecolor=green,colorlinks,bookmarks=true]{hyperref}
\setulcolor{blue} 

\begin{document}

\title{Hierarchical Disentanglement-Alignment Network for Robust SAR Vehicle Recognition}
\author{Weijie Li, Wei Yang${^*}$, Wenpeng Zhang, Tianpeng Liu${^*}$, Yongxiang Liu${^*}$, Li Liu
\thanks{This work was supported by the National Key Research and Development Program of China (No. 2021YFB3100800), the National Natural Science Foundation of China (No. 62376283, 61871384, 61901487, 61901498, and 61921001), and the Science and Technology Innovation Program of Hunan Province (No. 2022RC1092). ${^*}$Corresponding authors.}
\thanks{The authors are with the College of Electronic Science and Technology, National University of Defense Technology, Changsha, 410073, China (email: lwj2150508321@sina.com, yw850716@sina.com, zhangwenpeng08@nudt.edu.cn, everliutianpeng@sina.cn, lyx\_bible@sina.com, and dreamliu2010@gmail.com).}}

\markboth{This manuscript was submitted to IEEE Journal of Selected Topics in Applied Earth Observations and Remote Sensing}%
{Li \MakeLowercase{\textit{et al.}}: A Sample Article Using IEEEtran.cls for IEEE Journals}
\maketitle
\begin{abstract}

Vehicle recognition is a fundamental problem in SAR image interpretation. However, robustly recognizing vehicle targets is a challenging task in SAR due to the large intraclass variations and small interclass variations. Additionally, the lack of large datasets further complicates the task. Inspired by the analysis of target signature variations and deep learning explainability, this paper proposes a novel domain alignment framework named the Hierarchical Disentanglement-Alignment Network (HDANet) to achieve robustness under various operating conditions. Concisely, HDANet integrates feature disentanglement and alignment into a unified framework with three modules: domain data generation, multitask-assisted mask disentanglement, and domain alignment of target features. The first module generates diverse data for alignment, and three simple but effective data augmentation methods are designed to simulate target signature variations. The second module disentangles the target features from background clutter using the multitask-assisted mask to prevent clutter from interfering with subsequent alignment. The third module employs a contrastive loss for domain alignment to extract robust target features from generated diverse data and disentangled features. Lastly, the proposed method demonstrates impressive robustness across nine operating conditions in the MSTAR dataset, and extensive qualitative and quantitative analyses validate the effectiveness of our framework. 
\end{abstract}

\begin{IEEEkeywords}
Synthetic aperture radar (SAR), automatic target recognition (ATR), deep learning, domain alignment, robustness
\end{IEEEkeywords} 

\section{Introduction}
\label{Introduction}
\IEEEPARstart{T}{HANKS} to its attractive imaging capabilities in nearly all weather and illumination conditions, Synthetic Aperture Radar (SAR) has become an indispensable means of information acquisition in Earth observation. 
In recent years, SAR imaging techniques\cite{sun2021spaceborne, ref1} have been rapidly developing, and high-resolution SAR images can be accessed more easily than before, enabling a wide field of applications. As a result, the amount of SAR image data is growing rapidly, which requires the development of intelligent SAR image interpretation techniques. As a fundamental problem in SAR image interpretation, SAR vehicle recognition\cite{kechagias2021automatic,li2023comprehensive} aims to classify a vehicle into one of the predefined categories. It has various civilian and military applications, including transportation management, automatic driving, concealment detection, and military reconnaissance\cite{gagliardi2023satellite, rizzi2021navigation, frolind2011circular, zhang2021domain, 9915465}. Therefore, it has been an active area for several decades.

\begin{figure*}[!tb]
\centering
\includegraphics[width=18.1cm]{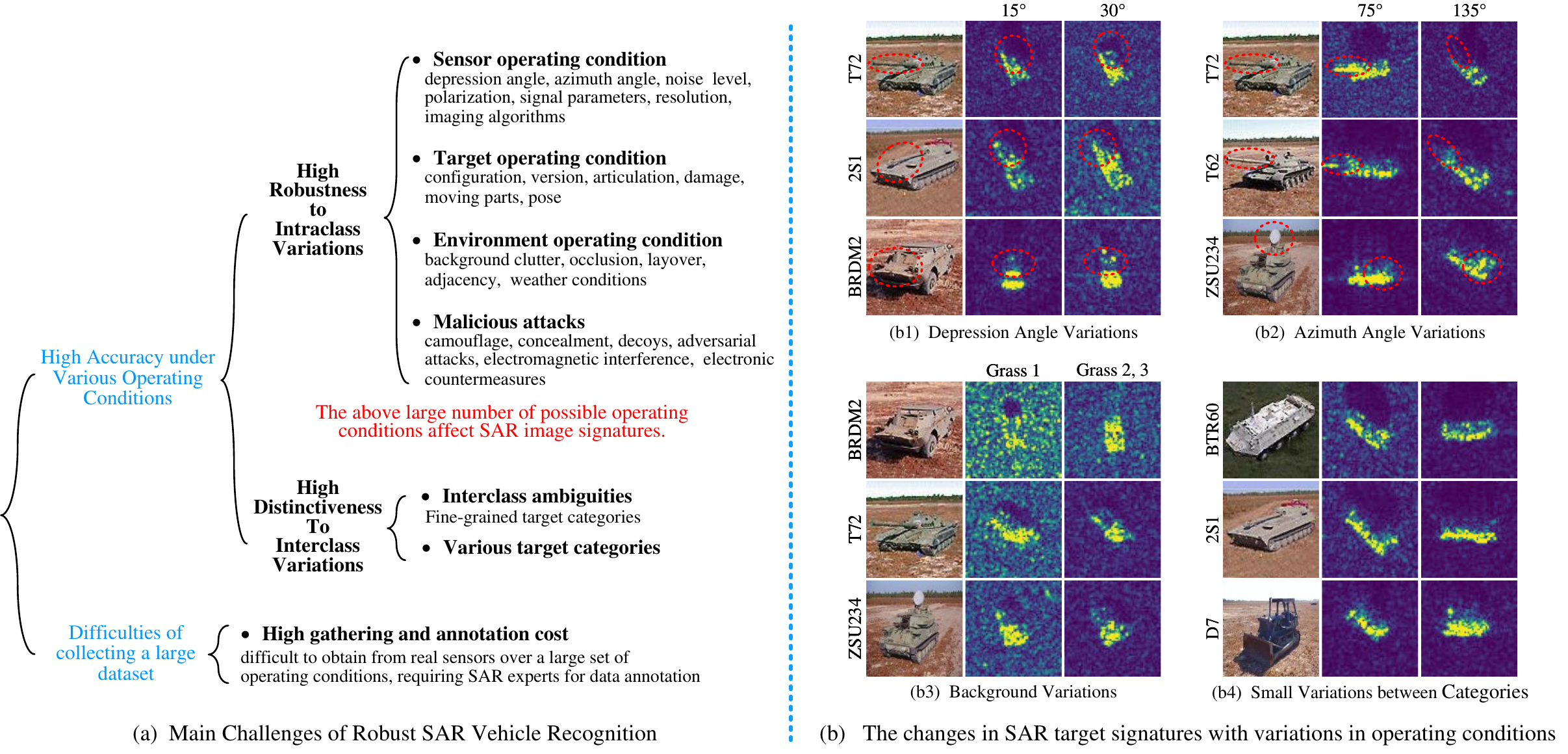}
\caption{
The main challenges of robust SAR vehicle recognition. Sub-figure \textbf{(a)} provides a taxonomy of these challenges brought by intraclass variations, interclass variations, and data collection. The right sub-figure illustrates typical variations with SAR images in the MSTAR dataset. On the right, \textbf{(b1)}, \textbf{(b2)}, and \textbf{(b3)} show large intraclass variations due to the sensitivity to operating conditions. \textbf{(b1)} contains target signatures and shadow variations in the red dashed line with different depression angles. \textbf{(b2)} displays the variation of target partial structures in the red dashed line with azimuth angles, such as the T-72 tank gun barrel is most visible in the vertical line-of-sight direction. \textbf{(b3)} illustrates that the intensity variation of different background clutter affects the adjacent target signatures. Therefore, SAR images of the same category have a large intraclass variation across operating conditions. In the end, sub-figure \textbf{(b4)} showcases small interclass variations between fine-grained vehicle target categories. Visual differences in SAR target signatures are much smaller than in natural images. The SAR images appear similar but are three different target categories.
}
\label{problem}
\end{figure*}

Through decades of effort, SAR Automatic Target Recognition (ATR) has witnessed significant progress. Especially in the past several years, deep learning has injected new vitality in this field and brought great success\cite{kechagias2021automatic, li2023comprehensive}. For instance, many existing methods have achieved over $99\%$ accuracy\cite{ref7, ref13, ref15, feng2022electromagnetic} on the widely used Moving and Stationary Target Acquisition and Recognition (MSTAR)\cite{ref4} dataset with ten categories of ground vehicles under standard conditions. Despite several decades' research in SAR ATR, most approaches have not been, however, capable of performing at a level sufficient for open, real-world applications\cite{kechagias2021automatic, li2023comprehensive, ref58,fan2019challenges}. Robust SAR vehicle recognition for practical applications is still far from being solved. \emph{What makes the problem of robust SAR vehicle recognition in the open world so challenging?} The main fundamental challenges of robust SAR vehicle recognition are summarized in Fig. \ref{problem} and discussed briefly in the following.

\IEEEpubidadjcol

\begin{itemize}
\item[$\bullet$] \textbf{High Robustness to Large Intraclass Variations.} As shown in Fig. \ref{problem}, challenges\cite{kechagias2021automatic, li2023comprehensive, ref58, fan2019challenges} in achieving high recognition accuracy stem from (1) the vast range of natural and adversary-induced difficult deployment conditions, and (2) the interclass ambiguities between potential fine-grained target categories. Natural intraclass variations include at least three types (see Fig. \ref{problem} for details): sensor operating conditions, target operating conditions, and environmental operating conditions. The scatter characteristics of the SAR target are highly sensitive to the aforementioned operating conditions, and therefore robust SAR ATR requires features that are highly robust to numerous possible operating conditions.
\item[$\bullet$] \textbf{High Distinctiveness to Small Interclass Variations.} The interclass ambiguities of some target categories, especially fine-grained target categories, demand great discrimination power from the features to distinguish between subtly different interclass variations.
\item[$\bullet$] \textbf{Lack of Large Target Datasets.} It is highly difficult to obtain SAR images over a large set of operating conditions from real sensors, and collecting and annotating large-scale SAR vehicle datasets is clearly more time-consuming and costly than in the natural images. Currently, methods for SAR vehicle recognition are mainly evaluated on the small MSTAR dataset collected under very constrained operating conditions. Many ATR methods have achieved near-perfect accuracy on MSTAR, suggesting the presence of strong bias in a small dataset. Therefore, the lack of large benchmark SAR vehicle datasets limits the power of deep learning methods requiring large amounts of training data and greatly impedes the development of SAR ATR technique.
\end{itemize} 

Early recognition systems utilized template matching\cite{ikeuchi1996invariant} and model-based\cite{diemunsch1998moving} methods, relying on numerous samples or accurate electromagnetic-scattering models. Another technique based on machine learning depended on expert-designed features and suitable classifiers\cite{li2023comprehensive}. In recent years, deep learning methods have shown their superiority over traditional methods by learning features in a data-driven manner\cite{kechagias2021automatic, li2023comprehensive, fan2019challenges}. However, the small dataset hampers the potential of deep learning, and many data augmentation methods\cite{ref17,ref19,ref23} have been employed to increase the richness of the dataset and enhance robustness to extra specific operating conditions. Based on data augmentation, domain alignment methods\cite{kwak2018speckle, he2021sar} were introduced to enhance features' invariance to intraclass variations further.

However, the potential of deep learning still needs to be more adequately exploited under the aforementioned fundamental challenges. 
Specifically, data biases such as background correlation\cite{kechagias2021automatic} in a small dataset interfere with domain alignment and data-driven models. This problem results in the model using background clutter for recognition and reduced robustness\cite{li2023discovering}. Training and test set shifts due to SAR imaging sensitivities are complex and do not satisfy the independent and identically distributed assumption. These issues need domain alignment to achieve robust recognition under different operating conditions on single-domain datasets (\emph{i.e.}, the training sets often contain a small number of operating conditions with data bias).

In this paper, to alleviate these challenges, a novel domain alignment framework named the Hierarchical Disentanglement-Alignment Network (HDANet) is proposed to achieve robust SAR vehicle recognition under various operating conditions. The novelty of our framework stands on careful consideration of the complex variation in target signatures and the clutter interference hidden by small data, and HDANet performs feature disentanglement and alignment through three steps.
Firstly, three data augmentation methods are designed considering complex variations in target local signatures, which generate diverse domain data for alignment.
Secondly, the multitask-assisted mask disentanglement module separates target and clutter regions at the feature layer because our previous work\cite{li2023discovering} on the explainability of deep learning illustrated that deep learning overfits background clutter to reduce training errors. To enhance the target region weight in the mask, the segmentation task and $l_1$ regularization are applied as auxiliaries to add target location and sparse constraints.
Thirdly, the robustness is increased under various operating conditions. We treat operating conditions as domains and improve robustness by performing domain alignment of target features. A contrastive loss and corresponding designed structures achieve domain alignment.
In the end, HDANet shows impressive robustness under MSTAR's nine operating conditions, and extensive qualitative and quantitative analyses demonstrate the effectiveness of our method. Moreover, the limitations of the proposed method are discussed.

The main contributions of this article are summarized as follows:

\begin{enumerate}
\item{A novel domain alignment framework named HDANet is proposed to achieve robust recognition under various operating conditions. To better evaluate robustness, we design nine operating conditions, including a scene variation that has not been previously discussed.}
\item{Inspired by analysis of deep learning explainability and target signature variations, HDANet integrates feature disentanglement and alignment into a unified framework with three significant modules. The disentanglement suppresses clutter and ensures correct feature representation, and the alignment further enhances the robustness of deep learning features.}
\item{Compared with existing attention mechanisms, HDANet achieves feature disentanglement under data bias by adding priori constraints and improving the computational approach. Compared with existing domain alignment methods, HDANet increases the diversity of domain data generation and considers the background clutter interference to achieve robust recognition under various operating conditions.}
\end{enumerate} 

The remainder of this paper is organized as follows. Sec. II introduces related studies in SAR vehicle recognition. Sec. III introduces the framework of HDANet. Sec. IV conducts extensive experiments to demonstrate the robustness of our method, and numerous qualitative and quantitative analyses discuss the advantages and limitations of our framework. Sec. V concludes the whole paper.

\section{Related Work}
This section reviews key issues in SAR ATR and focuses on robust recognition and deep learning-based approaches related to our work. Furthermore, considering our work draws on domain alignment, we briefly review this topic.

\subsection{Key Issues in SAR ATR}
Depending on the granularity of the target category, ATR can be divided into dimensions\cite{dudgeon1993overview} such as detection (distinguishing between targets and clutter), classification (determining the type of target, \emph{e.g.}, tank), and identification (determining the specific type of target, \emph{e.g.}, T72). Studies based on MSTAR focus on identification, which is the fine-grained classification of targets, and we review the related work on the robustness, accuracy, efficiency, and explainability of SAR ATR\cite{ref58, kechagias2021automatic,fan2019challenges} as follow:

\subsubsection{\textbf{Robustness}}
The main experimental settings for SAR vehicle target recognition based on the MSTAR dataset include Standard Operating Conditions (SOCs) and Extended Operating Conditions (EOCs)\cite{kechagias2021automatic}. The former is in a similar distribution and includes ten or three classes. The latter includes different variations for operating conditions. The popular EOC settings are the depression angle, configuration, and version variants in the \cite{ref7}. In addition, other articles add noise\cite{ref15,han2021synthetic}, occlusion\cite{feng2022electromagnetic}, and other variations to this benchmark. Variations in operating conditions lead to complex shifts in the distribution between training and test sets, which leads to the task of robust recognition, \emph{i.e.}, achieving a high and stable classification performance in these cases. Early SAR ATR systems included template-matching, model-based, and machine learning-based approaches\cite{kechagias2021automatic,fan2019challenges, li2023comprehensive}. In recent years, deep learning methods have also been widely applied in SAR vehicle recognition. 

\textbf{Template-matching methods.} Collecting many samples can create a template library, and recognition depends on designed features and matching criteria. 
Ikeuchi \emph{et al.}\cite{ikeuchi1996invariant} used deformable template matching based on the invariant histogram. 
Tan \emph{et al.}\cite{tan2019target} applied components of the target outline as matching templates. 

\textbf{Model-based methods} generate different images from 3D electromagnetic (EM) scattering or computer-aided design (CAD) models\cite{diemunsch1998moving}, and the recognition relies on accurate models and efficiency calculation.
Ma \emph{et al.}\cite{ma2016three} matched the EM model predicted scattering centers with the test image scattering points' location and intensity.
Ding\cite{ding2022model} designed three similarity degrees to synthetically evaluate the similarity between the test image and the 3D scattering center model.

\textbf{Machine learning-based methods} includes some key parts such as feature extraction and classifier. Previous work proposed many valuable methods including geometric structure features\cite{gao2010improved}, electromagnetic-scattering features\cite{li2019sar}, local descriptors\cite{dong2020keypoint}, and sparse representations\cite{hou2015sar}. Classifiers such as support vector machines and random forests have also been absorbed into SAR vehicle recognition\cite{li2023comprehensive}. And in recent years, deep learning-based methods has adopted an end-to-end approach to complete feature extraction and classification.

\textbf{Deep learning-based methods}
can better learn correlations in a dataset but require large amounts of diverse samples. For example, Chen \emph{et al.}\cite{ref7} proposed an all-convolutional network (A-ConvNet) with random cropping. A-ConvNet shows high accuracy under depression angle, configuration, and version variations but is not robust to random noise. Therefore, researchers are exploring various strategies to extracting robustness features. It is worth noting that the following methods can be combined to tackle the challenges in robust SAR vehicle recognition.
\begin{itemize}
\item[$\bullet$] 
\textbf{Data augmentation} is a technique used to increase the size and diversity of a training dataset artificially. Many data augmentation approaches have been used to enhance deep learning's robustness to specific operating conditions. Popular data augmentation in SAR vehicle recognition includes translation\cite{ref23}, random cropping\cite{ref7}, affine transformation\cite{ref26}, elastic distortion\cite{ref26}, power transformation\cite{ref17}, rotation\cite{ref17,ref23}, flipping\cite{ref17}, multiresolution\cite{yan2018convolutional}, occlusion\cite{yan2018convolutional}, and noise adding\cite{ref17,ref19,ref23,kwak2018speckle,yan2018convolutional}. By creating augmented images, the training dataset becomes larger and more diverse, helping the model better generalize and handle intraclass variations. Therefore, we designed three data augmentation methods to simulate partial changes and overcome the drawback of a small dataset.

\item[$\bullet$] 
\textbf{Clutter suppression.}
Due to background correlation interfering with the recognition, Zhou \emph{et al.}\cite{ref6} used masks to separate clutter and designed a large-margin softmax batch-normalization CNN. Heiligers \emph{et al.}\cite{ref25} employed CAD projection to separate target regions in input images for recognition and found that removing clutter in a similar scene may decrease the recognition rate. However, these methods are sensitive to hyperparameters or require the 3D structure of the target and imaging parameters. Some researchers integrated segmentation methods with deep learning optimization.
Li \emph{et al.}\cite{ref19} trained a fully connected codec with an input layer mask and sparse constraint to separate the target from the clutter. Ren \emph{et al.}\cite{ref13} proposed an Extended Convolutional Capsule Network (ECCNet) with a Convolutional Block Attention Module (CBAM) to suppress clutter. Moreover, channel attention was also used to address the effects of clutter\cite{ref9}. Wang \emph{et al.}\cite{wang2022sar} employed a multiscale CBAM for feature fusion. Therefore, the attention mechanism has been widely used in SAR target recognition. However, the question is whether the bias of a small dataset affects the attention mechanism. Our experiments found that CBAM enhances clutter in some MSTAR images to exploit background correlation better. Consequently, we use target location and sparse constraints to enhance the weight of target regions in masks.

\item[$\bullet$] 
\textbf{Feature extraction} methods that capture robust features with special structures are also being explored. Many works extract low-level and high-level features of vehicle targets. 
Lin \emph{et al.}\cite{lin2017deep} increased the depth and width of the model using two convolutional highway layers with different kernel sizes.
Shang \emph{et al.}\cite{ref8} developed a two-stage deep memory convolutional neural network to learn samples' spatial features.
Ai \emph{et al.}\cite{ai2021sar} used multi-kernel fusion to enhance the feature representation of Convolution Neural Network (CNN).
Deformable convolution kernel\cite{wang2020deformable,zhao2022attentional} was also applied to extract the scattering and morphological characteristics of the target. 
Inspired by the above work, we used multi-scale feature maps and capsule networks to enhance the feature representation of deep learning. The hybrid feature is another popular strategy using traditionally designed features to improve the robustness of auto-extracted deep learning features.
Zhang \emph{et al.}\cite{zhang2017multi} combined the designed multi-orientation spatial features with a bidirectional long short-term memory network. 
Zhang \emph{et al.}\cite{ref15} proposed a lightweight modified VGG16\cite{ref32} (MVGGNet) with pre-trained weights and combined this model with Attributed Scattering Centers (ASC). 
Feng \emph{et al.}\cite{feng2022electromagnetic} further used the physical features of ASC to constrain the deep learning features. 
Our approach addresses the shortcomings of deep learning features in SAR ATR, which can be used to improve the robustness of the deep learning modules in these hybrid methods.

\item[$\bullet$] 
\textbf{Transfer learning} has been applied to enhance robustness in small datasets due to the better diversity of large-scale datasets. In SAR vehicle recognition, transfer learning\cite{huang2019and} involves leveraging pre-trained models from other sensors or tasks, such as natural images or other remote sensing datasets, and adapting them to SAR vehicle recognition. The pre-trained models capture generic midlevel features from large datasets, which can be beneficial in addressing downstream tasks\cite{huang2019and}.

\item[$\bullet$] 
\textbf{Domain alignment.}
In SAR ATR, domain alignment addresses the differences between simulated and real data or extracts domain-invariant features for particular operating conditions. 
Wang \emph{et al.}\cite{wang2019sar} integrated meta-learning and adversarial learning for cross-domain and cross-task transfer learning from simulated to real data. 
Lewis \emph{et al.}\cite{lewis2019realistic} filled the gap between simulated and real data through generative adversarial networks. 
Kwak \emph{et al.}\cite{kwak2018speckle} proposed a Speckle-Noise-Invariant Network (SNINet) with $l_2$ regularization to align CNN feature maps after data augmentation. 
He \emph{et al.}\cite{he2021sar} applied a Task-Driven Domain Adaptation (TDDA) way to align the fully connected layer features of the simulated and real data by Multi-Kernel Maximum Mean Discrepancy (MK-MMD) for the large depression variation.
Two methods are similar to our work: SNINet\cite{kwak2018speckle} with $l_2$ contrastive loss and TDDA\cite{he2021sar} with MK-MDD. However, these domain alignment approaches do not consider clutter interfering with feature robustness. We draw lessons from traditional detection and recognition processes to disentangle the target and clutter at the feature layer before domain alignment and recognition. Moreover, we consider partial changes in target signatures under various operating conditions rather than invariance to specific operating conditions.
\end{itemize} 

\subsubsection{\textbf{Accuracy}}
In addition to a high accuracy rate under large intraclass variations, another high accuracy requirement is for small interclass variations. Therefore, in addition to high accuracy for existing fine-grained target categories, outlier rejection for various unknown target categories is also a concern task, including false alarm rate\cite{ref7} and open-set recognition\cite{ma2021open}. Chen \emph{et al.}\cite{ref7} set the confuser rejection rule of deep learning and tested operating characteristic curves with two confuser targets. Ma \emph{et al.}\cite{ma2021open} solved the open-set recognition problem by generative adversarial networks with classification and abnormal detection tasks.
\subsubsection{\textbf{Efficiency}}
SAR ATR systems need to consider development and deployment costs such as data collection, training costs, and hardware resources\cite{ref58, fan2019challenges, li2023comprehensive}. In addition to the data augmentation and transfer learning described above, methods for efficiency problems also include data generation\cite{luo2020synthetic}, electromagnetic simulation\cite{he2021sar,wang2019sar}, few-shot learning\cite{zhang2021domain}, and model compression\cite{zhang2020lossless}.
\subsubsection{\textbf{Explainability}}
Because of the black box problem, whether deep learning learns the correct feature representation in SAR ATR is an open question that deserves investigation. Previous explainability studies\cite{ref29,ref25,ref30,li2023discovering} used post-hoc methods to reveal the effect of target, clutter, and shadow regions on recognition, and the ablation studies\cite{ref30, ref6} discussed the influence of background clutter on the recognition rate. The above studies show that deep learning relies not only on target signatures but also on background clutter. This phenomenon is due to the bias in the small samples of the MSTAR dataset\cite{li2023discovering}. Since the data were collected under specific operating conditions, the background clutter and target categories are correlated. Therefore, we perform feature disentanglement to suppress background clutter and extract correct feature representations.

\subsection{Domain Alignment}
Since we consider operating conditions as domains, out-of-distribution generalization under different operating conditions is a domain generalization problem. Domain generalization\cite{zhou2022domain} aims to generalize out-of-distribution using only source data, while the target domain may be difficult to obtain or even unknown. Let $\mathcal{X}$ be the input (feature) space and $\mathcal{Y}$ the output (label) space. A domain $\mathcal{S}$ is associated with a joint distribution $P$ on $\mathcal{X}\times \mathcal{Y}$. Given $K$ train (source) domain $\mathcal{S}^{\mathrm{train}}= \left\{ \mathcal{S}_{k} \right\} ^{K}_{k=1}$ with different joint distributions $P_{k}\neq P_{k^{\prime}}$, $1\leq k\neq k^{\prime} \leq K$, the goal of domain generalization is to learn a model $f$ from $\mathcal{S}^{\mathrm{train}}$ to generalize on unseen test (target) domain $\mathcal{S}^{\mathrm{test}}$\cite{zhou2022domain} (\emph{i.e.}, $\mathcal{S}^{\mathrm{test}}$ is not included in $\mathcal{S}^{\mathrm{train}}$ and $P^{\mathrm{test}}\neq P^{\mathrm{train}}$):
\begin{equation}\label{eq_DG}
\min\limits_{f} \mathbb{E}_{(x, y)\in \mathcal{S}^{\mathrm{test}}} [\mathcal{L}(f(x), y)],
\end{equation}
where $\mathbb{E}$ is the expectation and $\mathcal{L}(\cdot,\cdot)$ is the loss function.

Domain alignment\cite{zhou2022domain} is one of the common methods to solve the domain generalization problem. This method minimizes differences between different source domains and extracts domain-invariant features: 
\begin{equation}\label{eq_DA}
\min\limits_{f} \mathbb{E}_{(x, y)\in \mathcal{S}^{\mathrm{train}}} [\mathcal{L}(f(x), y)+\lambda \mathcal{H}(P_{k}^{X}, P_{k^{\prime}}^{X})],
\end{equation}
where $\lambda$ is the tradeoff parameter to prevent a trivial solution, $\mathcal{H}(\cdot,\cdot)$ is domain distance metrics, $P^{X}$ is the marginal distribution. A common assumption is that the posterior distribution $P^{Y|X}$ remains stable, while domain shifts occur in the marginal distribution $P^{X}$. The domain distance measures\cite{zhou2022domain} include moments, contrastive loss, Kullback–Leibler divergence, Maximum Mean Discrepancy (MMD) distance, and adversarial learning. 
This paper assumes that domain shifts occur in $P^{X}$ and use the contrastive loss with cosine similarity. The different source domain data are generated by data augmentation.

Moreover, from a causal perspective\cite{ref67}, only the alignment of target features is meaningful. Mahajan \emph{et al.}\cite{mahajan2021domain} selected images of the same objects for domain generalization by a causal matching algorithm. Lv \emph{et al.}\cite{lv2022causality} designed a causality-inspired representation learning algorithm, including amplitude intervention, factorization, and adversarial mask modules to suppress non-causal factors such as background, style, and viewpoint. Our work is also inspired by the causal domain alignment method in computer vision and proposes feature disentanglement and alignment for SAR recognition. Due to target features changing with operating conditions, both causality and robustness of features are critical.

\section{Methodology}
This section introduces our method for robust SAR vehicle recognition. The main inspirations for designing a domain alignment framework for SAR are described in Sec. \ref{3-A}. The holistic framework of HDANet is present in Sec. \ref{3-B}, and its three modules are depicted in Sec. \ref{3-C}, \ref{3-D}, and \ref{3-E}, respectively.

\subsection{Motivation}
\label{3-A}
Feature disentanglement and alignment are our main concerns in adapting deep learning with domain knowledge for application to robust SAR vehicle recognition. 

\begin{figure}[!tb]
\centering
\includegraphics[]{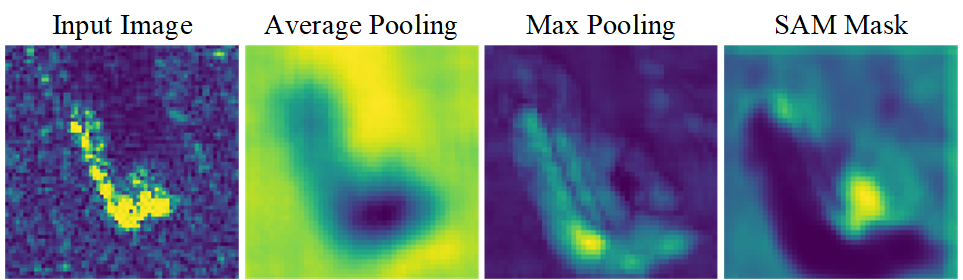}
\caption{Attention module results of CBAM. The Spatial Attention Mechanism (SAM) in CBAM generates masks based on the pooling results of feature maps. SAM mask shows that the background clutter has more weight than the target, which indicates that data bias can affect mask learning.}
\label{fig_atten}
\end{figure}

\textbf{Feature disentanglement.}
The training data in SAR vehicle recognition are collected under restricted conditions, and the clutter of each target class has different strengths\cite{li2023discovering}. So deep learning can use these differences to reduce training errors. Generally speaking, clutter amplitudes in diverse scenes obey different statistical distributions. The strong randomness of clutter can lead to instability of the extracted features. Therefore, the target and clutter features must be disentangled before domain alignment.
Motivated by the traditional Constant False Alarm Rate Detector (CFAR) algorithm\cite{rohling1983radar}, we can enhance the causality of a deep learning model in SAR with an input image mask to distinguish between targets and clutter. However, deep learning models may focus on the mask's hard edges\cite{ref25}. Moreover, the discontinuous strong clutter scattering point in SAR images increases learning difficulty. The input image mask learned by fully connected layers\cite{ref19} detected strong clutter points at a 5 dB signal-to-clutter ratio. Therefore, a soft constraint mask at the middle feature layer is more suitable for deep learning in SAR. A deep network can filter noise when compressing information\cite{ref64}, and middle-layer features are generic for different categories\cite{ref44}. Therefore, we extract masks at the feature layer to avoid noise points in the input image and to exploit the difference in middle patterns. This approach is similar to the attention mechanism.
Nevertheless, deep learning models cannot suppress the correlation of clutter in MSTAR without prior constraints. The attention module \cite{ref13} may overfit the background clutter, as shown in Fig. \ref{fig_atten}. Furthermore, its Sigmoid activation function maps the background region's smaller values to around 0.5. Besides, the image reconstruction task\cite{ref19,ref13} needs to reconstruct both the target and clutter in the SAR image, which enhances the impact of clutter as well. Therefore, implementing feature disentanglement in SAR vehicle recognition, \emph{i.e.}, separating target and background clutter, requires constraints on the mask to overcome the adverse effects of inconspicuous target signatures and a small dataset.

In addition to the mask, another common pre-processing method is center cropping to remove background clutter. This way requires a priori knowledge of the target size. Our analysis in Table \ref{tab_shapely} illustrates that even cropping to a quarter of the original image size does not eliminate background clutter interference because the center cropping retains the background clutter near the target.

\textbf{Feature alignment.}
Since SAR image properties are sensitive to operating conditions, the robustness of features under different operating conditions is critical for SAR target recognition. Extracting invariant target features under different operating conditions by domain alignment has a very attractive prospect for improving feature robustness. However, implementing domain alignment in SAR needs to address the following issues. Feature disentanglement solves the problem of extracting target features, and another question is how to generate domain data under different operating conditions. The training set in MSTAR only contains several constrained operating conditions (\emph{e.g.} different azimuth angles) due to the high cost of obtaining target data under various operating conditions. Therefore, data augmentation is applied to generate domain data effectively. The proposed domain data generation is designed based on the assumption of local perturbations in target signatures\cite{Ding2018research}. Specifically, the overall structure of the target does not change significantly with imaging parameters, but the position and magnitude of a few scattering points change. Therefore, various operating conditions lead to partial changes in the target pixel points of SAR images. Local perturbations of target signatures in different domains are assumed to be changes in image pixel points' position, relative value, and absolute value. Based on this assumption, we simulate different domain data with three data augmentation methods. In addition, the feature representation used for domain alignment needs to be modified according to SAR image properties.

Shortly, the feature disentanglement aims to suppress clutter and extract target features for alignment, and feature alignment extract invariant target features across operating conditions. Inspired by the above two principles, we designed a novel domain alignment framework based on SAR image properties to achieve robust recognition in various operating conditions.

\subsection{Overall Framework of HDANet}
\label{3-B}

\begin{figure*}[!tb]
\centering
\includegraphics[width=18.1cm]{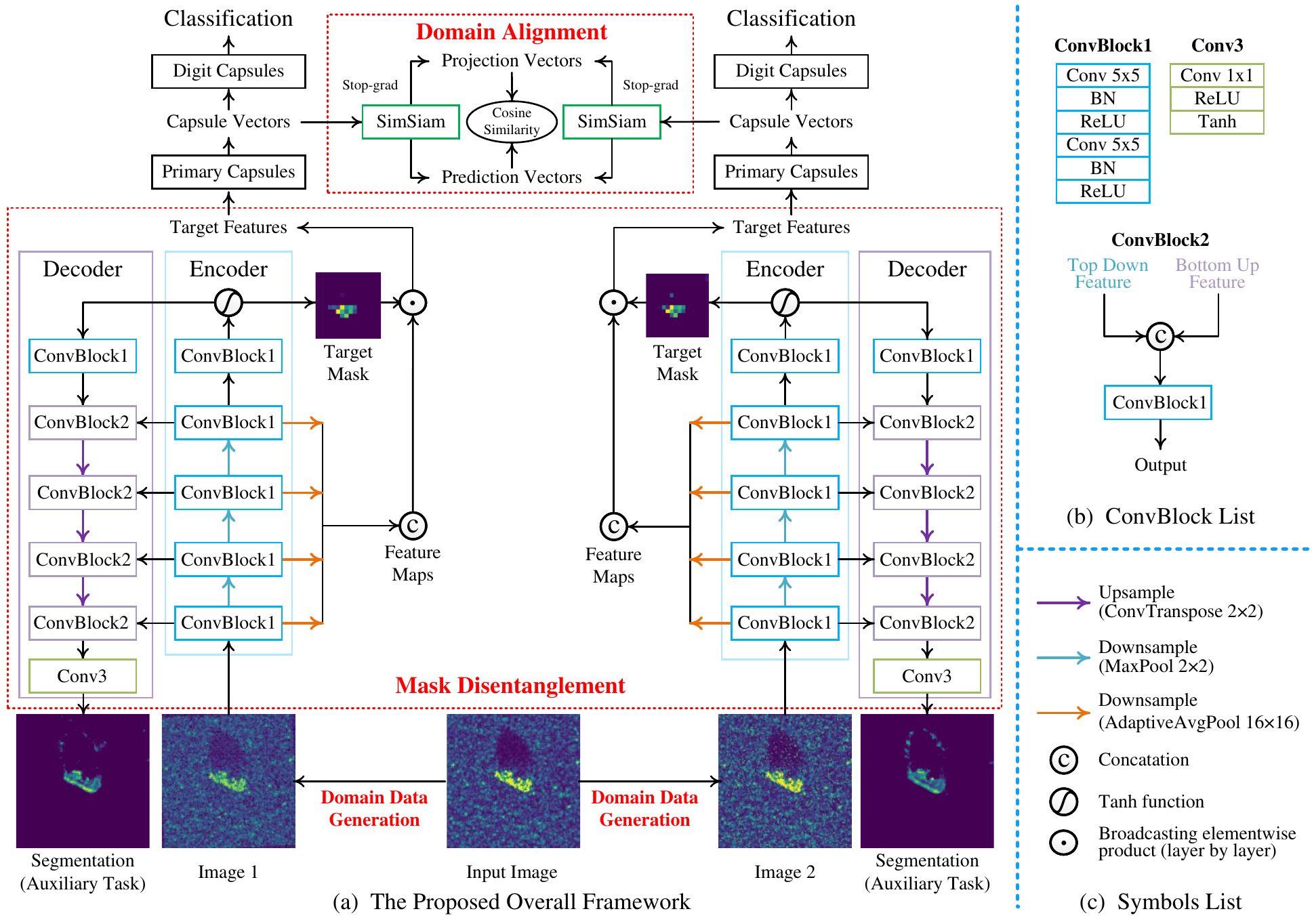}
\caption{The overall framework of HDANet. Sub-figure \textbf{(a)} provides its framework with three modules: domain data generation, multitask-assisted mask disentanglement, and domain alignment of target features. Three data augmentation methods are used in domain data generation. Mask disentanglement includes an encoder and decoder. The feature maps and the target mask are multiplied to get target features, and the disentangled target features are used for classification and domain alignment. We use the segmentation task and the $l_1$ loss to assist with the target mask. In the domain alignment module, target feature maps are converted to capsule vectors, and then the contrastive loss with cosine similarity, which SimSiam calculates, is used to enhance feature robustness. The parameters and structure are the same on both sides. The right sub-figure illustrates the details of \textbf{(b)} ConvBlocks (BN is batch normalization). and \textbf{(c)} Symbols list.}
\label{fig_framework}
\end{figure*}

The overall framework of our method is shown in Fig. \ref{fig_framework}, and HDANet has three modules to achieve robust recognition through feature disentanglement and alignment.

\textbf{Domain data generation} consists of three approaches: rotation, noise perturbation, and random replacement. Moreover, these augmentation methods are applied with different parameters and probabilities to simulate complex variations. A pair of images ($\boldsymbol{x}^{\prime}_1$ and $\boldsymbol{x}^{\prime}_2$) can be obtained by sampling the set $\mathcal{T}$ of data augmentation methods twice ($t_1$ and $t_2$) for domain alignment. This module generates diverse domain data from the MSATR single-domain training set.

\textbf{Multitask-assisted mask disentanglement} 
includes an encoder and a decoder. The encoder consisting of ConvBlock1 extracts the target mask $\boldsymbol{z}^{\mathrm{m}}$ and concatenated feature maps $\boldsymbol{z}^{\mathrm{c}}$, and the target mask and feature maps multiply to obtain the target features for alignment and recognition. The auxiliary multitask of the mask uses the segmentation task and sparse loss to suppress clutter effectively. The segmentation task is introduced by a decoder consisting of ConvBlock2 and Conv3 with automatic pseudo-labels and binary cross-entropy loss. And the sparsity constraint is added to the target mask with $l_1$ loss. The ConvBlock1, ConvBlock2, and Conv3 are depicted in Fig. \ref{fig_framework} (b). With extra prior constraints, this module can disentangle the target from clutter for subsequent domain alignment and recognition. 

\textbf{Domain Alignment of Target Features.} 
After disentangling the target from clutter, the model needs to be robust to the local perturbations in target features. We convert target feature maps to capsule vectors\cite{ref41}. The capsule vector $\boldsymbol{u}$ is applied as the final feature expression for domain alignment and classification because spatial relationships between features can be expressed through cosine similarity between capsule vectors. The domain alignment uses cosine similarity as the contrastive loss, and applying SimSiam\cite{ref46} mitigates the conflict between the contrastive loss and classification. Consequently, with a carefully designed domain alignment module further enhancing the invariance of capsule vectors, our framework achieves robust recognition under various operating conditions.

\begin{algorithm*}[tb]
\small
\caption{HDANet: Hierarchical Disentanglement-Alignment Network}%
\label{alg1}
\LinesNumbered
\KwIn{
a batch of $N$ data: images $\left\{ \boldsymbol{x}_{i} \right\} ^{N}_{i=1}$, classification labels $\left\{ y^{\mathrm{cls}}_{i} \right\} ^{N}_{i=1}$, and segmentation labels $\left\{ \boldsymbol{y}^{\mathrm{seg}}_{i} \right\} ^{N}_{i=1}$; 
models: parameters $\theta$, encoder$f^{\mathrm{enc}}_\theta$, decoder $f^{\mathrm{dec}}_\theta$, primary capsules $f^{\mathrm{pri}}_\theta$, digit capsules $f^{\mathrm{dig}}_\theta$, and SimSiam $f^{\mathrm{sim}}_\theta$; 
data augmentation set $\mathcal{T}$; optimizer; 
}
\KwOut{parameters $\theta$}
\small
\For{$\boldsymbol{x}_i\in \left\{ \boldsymbol{x}_{i} \right\}$, $y^{\mathrm{cls}}_{i}\in \left\{ y^{\mathrm{cls}}_i\right\}$, $\boldsymbol{y}^{\mathrm{seg}}_i\in \left\{ \boldsymbol{y}^{\mathrm{seg}}_i\right\}$}
{
$t_1\in \mathcal{T}$ and $t_2\in \mathcal{T}$ 
\tcp*{sample rand data augmentation}
$\boldsymbol{z}^{\mathrm{c}}_1, \boldsymbol{z}^{\mathrm{m}}_1 \gets f^{\mathrm{enc}}_{\theta}(t_1 (\boldsymbol{x}_i))$ and $\boldsymbol{z}^{\mathrm{c}}_2, \boldsymbol{z}^{\mathrm{m}}_2 \gets f^{\mathrm{enc}}_{\theta}(t_2 (\boldsymbol{x}_i))$ 
\tcp*{compute feature maps and target mask}
$\boldsymbol{o}_1 \gets f^{\mathrm{dec}}_{\theta}(\boldsymbol{z}^{\mathrm{c}}_1, \boldsymbol{z}^{\mathrm{m}}_1)$ and $\boldsymbol{o}_2 \gets f^{\mathrm{dec}}_{\theta}(\boldsymbol{z}^{\mathrm{c}}_2, \boldsymbol{z}^{\mathrm{m}}_2)$ 
\tcp*{compute segmentation results}
$\boldsymbol{u}_1 \gets f^{\mathrm{pri}}_{\theta}(\boldsymbol{z}^{\mathrm{c}}_1\otimes \boldsymbol{z}^{\mathrm{m}}_1)$ and $\boldsymbol{u}_2 \gets f^{\mathrm{pri}}_{\theta}(\boldsymbol{z}^{\mathrm{c}}_2\otimes \boldsymbol{z}^{\mathrm{m}}_2)$ 
\tcp*{disentangle and convert to vectors}
$\boldsymbol{m}_1, \boldsymbol{n}_1 \gets f^{\mathrm{sim}}_\theta(\boldsymbol{u}_1)$ and $\boldsymbol{m}_2, \boldsymbol{n}_2 \gets f^{\mathrm{sim}}_\theta(\boldsymbol{u}_2)$
\tcp*{compute projection and prediction vector}
$\boldsymbol{v}_1 \gets f^{\mathrm{dig}}_\theta(\boldsymbol{u}_1)$ and $\boldsymbol{v}_2 \gets f^{\mathrm{dig}}_\theta(\boldsymbol{u}_2)$
\tcp*{compute classification vectors}
$l^{\mathrm{cls}}_{i} = \sum_{j}^{1,2} l^{\mathrm{mar}}(\boldsymbol{v}_j, y^{\mathrm{cls}}_i)$
\tcp*{compute classification loss}
$l^{\mathrm{con}}_{i} \gets 1-\frac{1}{2}\frac{\mathrm{stopgrad}(\boldsymbol{m}_1)}{\Vert \mathrm{stopgrad}(\boldsymbol{m}_1)\Vert_2}\cdot \frac{\boldsymbol{n}_2}{\Vert \boldsymbol{n}_2\Vert_2}-\frac{1}{2}\frac{\mathrm{stopgrad}(\boldsymbol{m}_2)}{\Vert \mathrm{stopgrad}(\boldsymbol{m}_2)\Vert_2}\cdot \frac{\boldsymbol{n}_1}{\Vert \boldsymbol{n}_1\Vert_2}$
\tcp*{compute contrastive loss}
$l^{\mathrm{seg}}_{i} \gets \sum_{j}^{1,2} \mathrm{binary\ cross}\mbox{-}\mathrm{entropy}(\boldsymbol{o}_{j}, \boldsymbol{y}^{\mathrm{seg}}_{i})$
\tcp*{compute segmentation loss}
$l^{\mathrm{spa}}_{i} \gets \sum_{j}^{1,2} \Vert \boldsymbol{z}^{\mathrm{m}}_j \Vert_1$
\tcp*{compute sparse loss with mean $l_1$ loss}
$\mathcal{L}_i \gets l^{\mathrm{cls}}_{i}+l^{\mathrm{con}}_{i}+\alpha \cdot l^{\mathrm{seg}}_{i}+\beta \cdot l^{\mathrm{spa}}_{i}$ 
\tcp*{compute the loss for $\boldsymbol{x}_i$}
}
$\delta \theta \gets \frac{1}{N}\sum^N_{i=1} \partial_{\theta}\mathcal{L}_i$

$\theta \gets \mathrm{optimizer}(\theta, \delta \theta)$
\tcp*{update parameters}
\end{algorithm*}

For clarity, we briefly describe the training process of HDANet in Algorithm \ref{alg1}. The total loss $\mathcal{L}$ is in \ref{eq3}, including the classification loss $l^{\mathrm{cls}}$ of the capsule vectors, the contrastive loss $l^{\mathrm{con}}$ with the cosine similarity, the segmentation loss $l^{\mathrm{seg}}$ with binary cross-entropy, and the sparse constraint $l^{\mathrm{spa}}$ with $l_{1}$ regularization of the target mask. In the test phase, the domain data generation is removed, and the classification result is obtained by inputting a SAR image. 
\begin{equation}
\label{eq3}
\mathcal{L} = l^{\mathrm{cls}}+l^{\mathrm{con}}+\alpha \cdot l^{\mathrm{seg}}+\beta \cdot l^{\mathrm{spa}}
\end{equation}

\subsection{Domain Data Generation}
\label{3-C}
Since clutter in different scenes obeys various distributions without stable correlation with target categories, we use data augmentation to simulate target signatures rather than background clutter variations. For simplicity, we apply data augmentation to the whole image area and use the mask to separate the target region so that the clutter region does not affect the discrimination and robustness of target features. The domain data generation module has three methods to simulate the variations in SAR image pixels: position, relative value, and absolute value. 

\textbf{Rotation.} The position is simulated by rotating at different angles $\theta$. It is worth mentioning that SAR imaging causes target shadows in the line-of-sight direction and the shadow above the target in the MSTAR dataset. The amplitude of the scattering point remains stable in real situations only under small viewing angle changes as well. Therefore the rotation operation is performed within a small angle range to maintain this imaging relationship. 

\textbf{Noise Perturbation.} The shift degree of the target's strong and weak scattering points is not the same across operating conditions, so noise perturbation adds Gaussian white noise $N(\mu, \sigma)$ to the original image to simulate relative value changes:
\begin{equation}
\label{eq_np1}
x^{\prime}_{n}=x_{n}+x^{\mathrm{wgn}}, x^{\mathrm{wgn}}\sim A\cdot N(\mu, \sigma),
\end{equation}
where $x^{\prime}_{n}$ is the $n$-th pixel point after augmentation, $x_{n}$ is the $n$-th pixel point before augmentation, $j$ iterates over all pixel points of the image, $A$, $\mu$, $\sigma$ are the magnitude, mean and standard deviation of the Gaussian distribution respectively.

\textbf{Random replacement} simulates absolute value changes by replacing a different proportion $p$ of pixel points with a uniform distribution $U(0, 1)$:
\begin{equation}
\label{eq_np2}
x^{\prime}_{n}=x^{\mathrm{uni}}, x^{\mathrm{uni}}\sim U(0, 1),
\end{equation}
where $n$ iterates over the $p$-proportion pixel points of the image. Since the single point's amplitude in a SAR image is unstable across operating conditions, the random replacement prevents the model from being sensitive to the intensity of a single pixel point.

Based on the domain data generation consisting of the above three data augmentation methods, we generate a pair of image pairs ($x^{\prime}_1$ and $x^{\prime}_2$) from the input image $x$ for domain alignment.

\subsection{Multitask-Assisted Mask Disentanglement}
\label{3-D}
This module includes an encoder and a decoder. The codec skeleton is the U-shaped structure of U-Net\cite{ref33}, commonly used in small sample tasks. As shown in Fig. \ref{fig_framework}, different modules are variants of the U-net basic module, and we tuned the layers and parameters of the original U-Net to fit the picture size of the MSTAR dataset. In the encoder, a soft target mask $\boldsymbol{z}^{\mathrm{m}}$ is extracted at the middle layer and multiplied with concatenated feature maps $\boldsymbol{z}^{\mathrm{c}}$ that have multi-scale details. The decoder is used to perform the segmentation task. Then, we introduce the multitask setting of this module. In order to successfully learn the target mask in a small SAR dataset, the segmentation task and sparsity constraint are auxiliary tasks to introduce position and sparse priors. 

\textbf{Segmentation task} provides information on the target location compared to the reconstruction task, making it necessary for the model to distinguish between the spatial regions of the target and clutter. We use the decoder to apply the auxiliary segmentation task and increase the discrepancy between targets and clutter in different layers through the U-shaped structure. The segmentation loss $l^{\mathrm{seg}}$ is the binary cross-entropy, and pseudo-labels $\boldsymbol{y}^{\mathrm{seg}}$ are multiple class saliency maps\footnote{Multiple class saliency map is a variant of Grad-Cam\cite{ref49} to eliminate category information.}\cite{fu2019multicam} of the pre-trained model VGG16\cite{ref32}. Moreover, we use SmoothGrad\cite{ref51} to average saliency maps and remove strong clutter points. To achieve a balance between accuracy and efficiency, we design this method of automatically generating segmentation pseudo-labels, but other ways to generate pseudo-labels are also feasible to provide target region information. 

\textbf{Sparsity constraint} of the target mask is due to the sparsity property of the target pixel compared to the whole image pixel. In the MSTAR dataset, the vehicle range in size from 4.1 m to 9.5 m long and from 2.3 m to 3.6 m wide, and the image is 128 × 128 pixels (38.4 m × 38.4 m). Moreover, the target region is small under airborne or satellite-based SAR platforms. The $l_1$ constraint is added to the target mask as the sparse loss $l^{\mathrm{spa}}$ to help remove clutter further. Moreover, we use ReLU to generate a truncated Tanh activation function so that the zero value of the background region maps to zero instead of 0.5 in the Sigmoid function. 

We achieve feature disentanglement by the target mask with the above two auxiliary tasks, and this module enhances feature robustness by preventing clutter from being used for recognition and domain alignment.

\subsection{Domain Alignment of Target Features}
\label{3-E}
Multitask-assisted mask disentanglement module extracts target features and suppresses clutter. However, not all target features are robust under different SAR operating conditions. Therefore, we use domain alignment of target features to extract invariant features. 
As shown in Fig. \ref{fig_framework}, the target feature maps are converted to capsule vectors for classification and domain alignment. Specifically, we consider the robust feature representation (\emph{e.g.}, capsule vectors) with domain alignment in SAR target recognition. 

\textbf{Feature representation.} The previous domain alignment methods used CNN feature maps\cite{kwak2018speckle} or Fully Connected (FC) layer features\cite{he2021sar}. Other work showed that using capsule vector is more robust in extended operating conditions\cite{ref13} than CNN and FC. Considering the target's overall structural information is more stable under the local target perturbation, we use the capsule vector \cite{ref41,ref13} to preserve the spatial information between features. The target feature map is converted to a capsule vector $\boldsymbol{u}$ whose magnitude represents the probability and direction represents spatial information (\emph{i.e.}, cosine similarity between features). The digit capsules perform recognition based on the magnitude and direction of the classification vectors $v$ in Fig. \ref{fig_framework}. The classification loss $l^{\mathrm{cls}}$ is the margin loss $l^{\mathrm{mar}}$ in \cite{ref41} with consistent hyperparameters ($w^{+}=0.9, w^{-}=0.1$, $\eta=0.5$): 
\begin{equation}
\begin{split}
l^{\mathrm{mar}} = & \sum_k T_{k}\cdot \mathrm{max}(0,w^{+}-\Vert \boldsymbol{v}_{j,k}\Vert)^2 \\
& +\eta (1-T_{k})\cdot \mathrm{max}(0,\Vert \boldsymbol{v}_{j,k}-w^{-}\Vert)^2, \\
\end{split}
\end{equation}
\begin{equation}
l^{\mathrm{cls}} = \sum_{j}^{1,2} l^{\mathrm{mar}}(\boldsymbol{v}_j, y^{\mathrm{cls}})
\end{equation}
where the first term of $l^{\mathrm{mar}}$ encourages correct prediction probability over $w^{+}$, the second term of $l^{\mathrm{mar}}$ penalizes incorrect prediction probability higher than $w^{-}$, $\boldsymbol{v}_{j,k}$ is the $k$-th classification vector of $\boldsymbol{v}_{j}$, $T_{k}=1$ if the classification label $y^{\mathrm{cls}}$ is class $k$, and $\eta$ prevents the initial learning from shrinking the magnitude of all digit capsules.

\textbf{Domain alignment.} Correspondingly, the domain alignment module uses the cosine similarity as contrastive loss and SimSiam\cite{ref46} structure. According to the SimSiam structure, the contrastive loss $l^{\mathrm{con}}$ is shown below:
\begin{equation}
\label{eq1}
D(\boldsymbol{m},\boldsymbol{n})=\frac{1}{2}-\frac{1}{2}\frac{\boldsymbol{m}}{\Vert \boldsymbol{m}\Vert_2}\cdot \frac{\boldsymbol{n}}{\Vert \boldsymbol{n}\Vert_2},
\end{equation}
\begin{equation}
\label{eq2}
l^{\mathrm{con}}=D(\mathrm{stopgrad}(\boldsymbol{m}_1),\boldsymbol{n}_2)+D(\mathrm{stopgrad}(\boldsymbol{m}_2),\boldsymbol{n}_1),
\end{equation}
where $\boldsymbol{m}$ and $\boldsymbol{n}$ are the projection and prediction vectors in Fig. \ref{fig_framework} respectively. The stopping gradient and asymmetric structure of SimSiam increase the interclass distance and avoid trivial constant solutions impairing discrimination. Other methods used the hyperparameter tradeoff\cite{kwak2018speckle, he2021sar} or negative samples\cite{zhou2022domain} to solve the problem of identical representation in domain alignment.

As discussed above, we designed a robust feature representation with domain alignment. Eventually, with the aforementioned domain data generation and mask disentanglement modules, robust features under various operating conditions are obtained by domain alignment of target features. In the proposed integrated framework of feature disentanglement and alignment, we mainly address the two problems in SAR vehicle recognition methods: the existing mask methods overfit the background clutter, and the current domain alignment methods achieve recognition for several specific operating conditions.

\section{Experiments}
\label{Experiments}
In this section, we evaluated the robustness of the proposed method on the MSTAR dataset and analyzed the strengths and weaknesses of HDANet. We first described the experimental setting and implementation details in Sec. \ref{Experimental Setting} and \ref{Implementation details}. The robustness of HDANet is evaluated in the MSTAR dataset's nine operating conditions compared to other methods in Sec. \ref{Results}. We then performed extensive quantitative and qualitative analyses in Sec. \ref{Analysis} to demonstrate the effectiveness of our method. Ultimately, we explored the limitations in Sec. \ref{Limitations}.

\subsection{Dataset and Experimental Settings}
\label{Experimental Setting}
\textbf{Dataset description.}
Sandia National Laboratory collected and released the popular MSTAR dataset\cite{ref4} with a 10-GHz X-band spotlight SAR sensor. This dataset contains ten categories of ground military targets: infantry vehicle (BMP2), patrol car (BRDM2), personnel carrier (BTR60, BTR70), tank (T62, T72), howitzer (2S1), bulldozer (D7), truck (ZIL131), and anti-aircraft (ZSU234). The resolution of each SAR image is 0.3 × 0.3 m, and MSTAR data are acquired at full azimuth angles from 0° to 360°. However, only some targets have different depression angles (\emph{e.g.}, 15°, 17°, and 30°) and scenes (grasslands in New Mexico, northern Florida, and northern Alabama\cite{ref58}). We used the official tool to convert original SAR data into JPEG format with 128 × 128 pixels by linear transformation and automatic contrast enhancement. 

\textbf{Experimental settings.}
\begin{table*}[!tb]
\centering
\caption{Experimental Setting under Standard Operating Condition (SOC) and Extended Operating Conditions (EOCs) of MSTAR. \\Robustness to EOCs Includes Sensor (Depression Angle, Azimuth Angle, and Noise Level), \\Target (Configuration and Version), and Environment (Occlusion and Scene)}
\label{table_setting}
\renewcommand\arraystretch{1.25}
\begin{tabular}{lllll} 
\toprule
\multicolumn{1}{c|}{\multirow{2}{*}{Operating conditions}} & \multicolumn{4}{l}{Setting of training set (test set)} \\
\cline{2-5}
\multicolumn{1}{c|}{} & \# Categories & \# Samples in total & Depression angle & Scene \\ 
\cmidrule(){1-5}
SOC & 10 (10) & 3671 (3203) & 17° (15°) & \multirow{8}{*}{Grass 1 \& HB (Grass 1 \& HB)}\\
EOC-Depression & 4 (4) & 1195 (1151) & 17° (30°) & \\
EOC-Azimuth & 10 (10) & downsampling of 3671 (3203) & 17° (15°) & \\
EOC-Gaussian & 10 (10) & 3671 (3203 with Gaussian noise) & 17° (15°) & \\
EOC-Random & 10 (10) & 3671 (3203 with random noise) & 17° (15°) & \\
EOC-Configuration & 4 (5 variants) & 996 (2710) & 17° (17° \& 15°) & \\
EOC-Version & 4 (7 variants) & 996 (3569) & 17° (17° \& 15°) & \\
EOC-Occlusion & 10 (10) & 3671 (3203 with occlusion) & 17° (15°) & \\
EOC-Scene & 3 (3) & 1772 (743) & 45° \& 30° (45° \& 30°) & Grass 1 (Grass 2 \& 3) \\
\bottomrule
\multicolumn{5}{l}{\begin{tabular}[c]{@{}l@{}}Note: Different parameters of EOC-Azimuth/Gaussian/Random/Occlusion control the level of variations compared to SOC (see Sec. \ref{Experimental Setting} \\for detailed presentation). Detailed information is available in the Supplementary Materials and on our GitHub.\end{tabular}}
\end{tabular}
\end{table*}
We discussed nine operating conditions based on the MSTAR dataset in Table \ref{table_setting} to evaluate the performance of our method comprehensively. The standard operating condition means that the imaging conditions of the training and test sets are similar. In extended operating conditions, different operating conditions lead to complex variations in target signature and background clutter\cite{ref58, fan2019challenges}. We built an extensive EOC setting, including sensor (depression angle, azimuth angle, and noise level), target (configuration and version), and environment (occlusion and scene). The EOC-Gaussion/Random/Occlusion are simulation settings, and others are measured data. The detailed settings are discussed below:

\subsubsection{\textbf{SOC (standard operating condition)}} 
The difference between the training and test sets is minor under SOC, and the main challenge is the small interclass differences. As shown in Table \ref{table_setting}, the training set's depression angle under SOC is 17°, and the test set is 15°. Ten categories of targets include BMP2 (C21, 9563, 9566), BRDM2, BTR60, BTR70, T62, T72 (132, 812, S7), 2S1, D7, ZIL131, and ZSU234.

\subsubsection{\textbf{EOC-Depression (depression angle variation)}} 
The intensity of pixel points in SAR images is related to the depression angle, and a large depression angle can change target signatures and enhance background clutter. Therefore, robustness to depression angle variation is critical to the sensor setup. Following the setting in \cite{ref7}, four targets are selected to discuss the performance from 17° to 30° in Table \ref{table_setting}. These targets include BRDM2, T72 (A64), 2S1, and ZSU234. 

\subsubsection{\textbf{EOC-Azimuth (azimuth angle variation)}} 
Due to the anisotropic scattering of the different structures in targets, local signatures vary at different azimuth angles. Since the SAR images of the MSTAR dataset are collected at different azimuth angles, the robustness to azimuth angle variation is tested by reducing training data from 90\% to 10\% under SOC in Table \ref{table_setting}.

\subsubsection{\textbf{EOC-Gaussion (Gaussian noise corruption)}} 
Sensor noise can significantly affect the properties of the measured SAR images, but additive noise in MSTAR is below -30 dB, which is far from realistic situations. As shown in Table \ref{table_setting}, the Signal-to-Noise Ratio (SNR) of additive Gaussian white noise\cite{ref15,han2021synthetic} is from 10 dB to -10 dB based on the SOC setting. 

\subsubsection{\textbf{EOC-Random (random noise corruption)}} 
Another noise setting is randomly replacing the original pixel values with a uniform distribution noise\cite{ref7, tan2019target}. This approach simulates the degree of random strong clutter point interference. Based on the SOC setting, random noise in Table \ref{table_setting} has a replacement rate of 5\% to 25\%. 

\subsubsection{\textbf{EOC-Configuration (configuration variant)}} 
The vehicle configurations often change depending on different actual needs, such as fuel containers and other accessories fixed to the vehicle. Following the setting in \cite{ref7}, the configuration variant has BMP2 (9563), BRDM2, BTR70, and T72 (132) as traing set, and T72 (A32, A62, A63, A64, S7) as 5 test variants. 

\subsubsection{\textbf{EOC-Version (version variant)}} 
Similar to the configuration variant, various vehicle versions are produced for different needs with a similar global structure and different local details. Following the setting in \cite{ref7}, the configuration variant has BMP2 (9563), BRDM2, BTR70, and T72 (132) as the training set. The BMP2 (9566, C21) and T72 (A04, A05, A07, A10, 812) are 7 test variants. 

\subsubsection{\textbf{EOC-Occlusion (occlusion interference)}} 
Objects between the target and sensor, such as trees and buildings, can obscure or attenuate the signal strength reflected from the target. Therefore, occlusion can severely eliminate or diminish some of the target signatures. According to \cite{feng2022electromagnetic}, the square of different pixel sizes (5 × 5, 10 × 10, and 15 × 15) in the target and the adjacent area (64 × 64 at picture center) are randomly set to zero in Table \ref{table_setting}.

\subsubsection{\textbf{EOC-Scene (scene variation)}} 
Background clutter in different scenes can vary significantly and affect the scattering points of adjacent target parts. Although the scenes in MSTAR are all flat grassland from different locations, the grass's height, sparsity, and water content affect its electromagnetic scattering\cite{ulaby2019handbook, Huang2017Studies}. BRDM2, T72 (A64), and ZSU234 in Grass 1 are the training set. ZSU234 in Grass 2 and BRDM2 and T72 (A64) in Grass 3 are the test data in Table \ref{table_setting}.

\subsection{Implementation details} 
\label{Implementation details}
\begin{table}[!tb]
\centering
\caption{Data Augmentation Details}
\label{data_aug}
\renewcommand\arraystretch{1.25}
\begin{tabular}{ccc} \toprule
Approach & Parameter & Probability \\ \midrule
Rotation & $\theta \sim U(-5^\circ, 5^\circ)$ & 0.3 \\
Noise Perturbation & $A\cdot N(0.1, 0.1), A\sim U(0.5, 1.5)$ & 0.2 \\
Random Replacement & $p\sim U(0, 5\%)$ & 0.2 \\ \bottomrule
\end{tabular}
\end{table}
\begin{figure}[!tb]
\centering
\includegraphics[width=8.8cm]{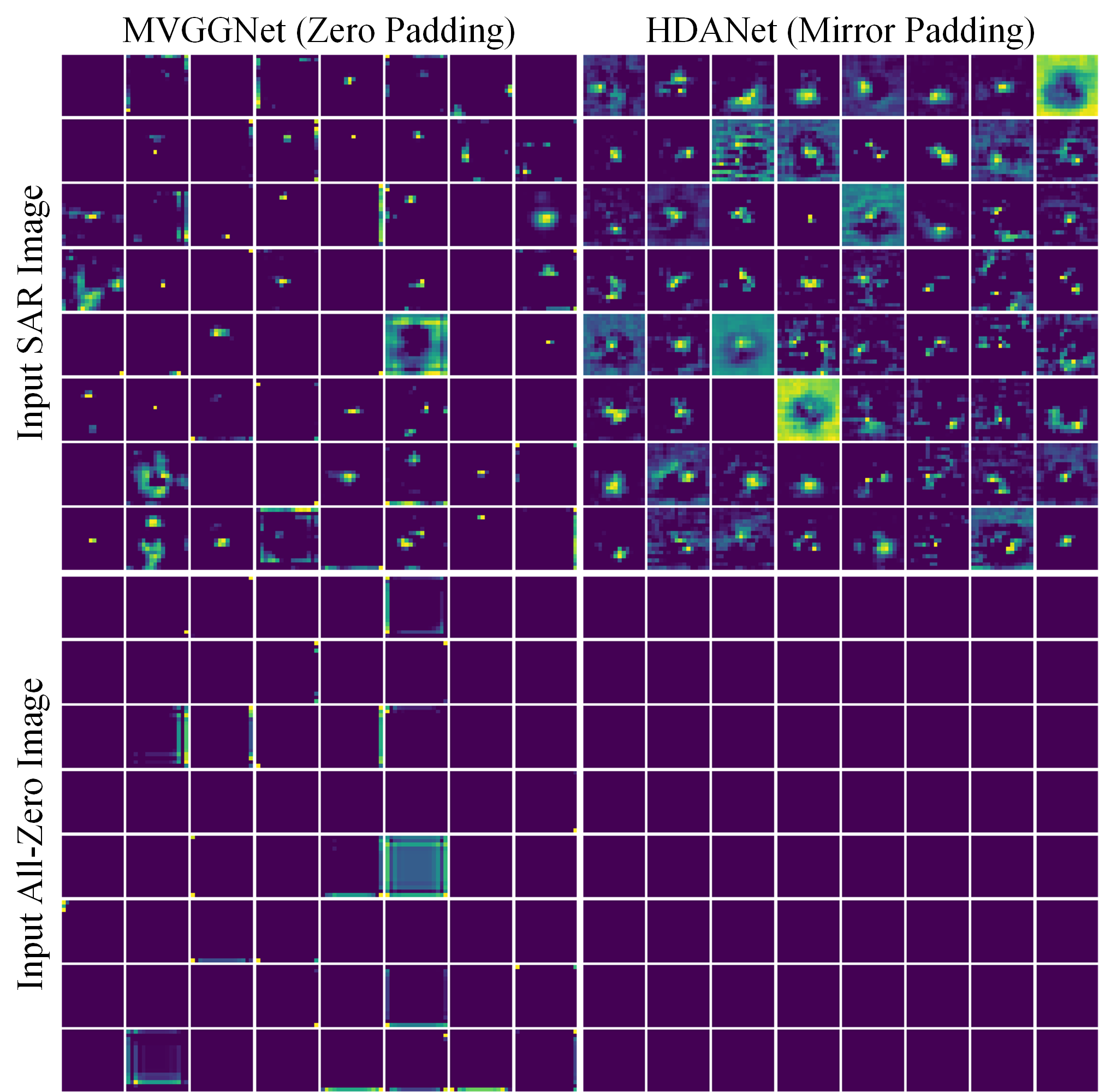}
\caption{Feature maps of different padding methods (feature maps on the left are MVGGNet\cite{ref15} with zero padding, and on the right are feature maps of HDANet (ours) with mirror padding; input from the top is an MSTAR image, and from the bottom is an all-zero image). Models with zero padding produce target-independent edge and center artifacts in the left feature maps\cite{ref48}.}
\label{result_padding}
\end{figure}

\textbf{Hyperparameter settings.}
Data augmentation details in this paper are shown in Table \ref{data_aug}. The data augmentation methods used different probabilities so that the results have a 50\% of being the original image. Noise perturbation was adjusted to $A\cdot N(0, 0.8)$, $A\sim U(0, 1)$ under EOC-Gaussion. The loss hyperparameter $\alpha$ was 1e-1 as an auxiliary task, and $\beta$ was 1e-2 to control mask sparsity. NAdam algorithm\cite{opt} was employed as the optimizer with an initial learning rate of 3e-4, weight decay of 5e-4, and an exponential learning rate decline of 0.98. The batch size was 64, and the epoch was 100. Notably, we used mirror padding instead of zero padding to eliminate feature-map artifacts in Fig. \ref{result_padding}. Our code is available at \url{https://github.com/waterdisappear/SAR-ATR-HDANet}.

\textbf{Compared methods.}
We used A-ConvNet\cite{ref7}, ECCNet\cite{ref13}, MVGGNet\cite{ref15}, SNINet\cite{kwak2018speckle} and TDDA\cite{he2021sar} as compared methods. A-ConvNet is one of the classical deep learning methods successfully applied to recognizing SAR vehicles. ECCNet is a method that uses the capsule network and the attention module. We combined CFAR with them to verify the effectiveness of traditional mask methods. MVGGNet with pre-training weight has good feature extraction ability by transfer learning. SNINet and TDDA are similar domain alignment methods in SAR vehicle recognition. All methods used the same data augmentation methods. Five repetitions of experiments were used to calculate the overall accuracy (OA) and the standard deviation (STD).

\subsection{Results under SOC and EOCs}
\label{Results}

\begin{table*}[tb]
\centering
\caption{Performances of Different Methods under Different Operating Conditions}
\label{table_result}
 \renewcommand\arraystretch{1.25}
\begin{tabular}{cccccccccc} 
\toprule
\multicolumn{1}{c|}{\multirow{2}{*}{Method}} & \multirow{2}{*}{SOC} & \multicolumn{8}{c}{EOC} \\ 
\cline{3-10}
\multicolumn{1}{c|}{} & & Depression & Azimuth & Gaussian & Random & Configuration & Version & Occlusion & Scene \\ 
\cmidrule(){1-10}
A-ConvNet & 98.12±0.23 & 89.50±1.46 & 63.95+4.20 & 71.36±1.11 & 91.82±1.03 & 96.69±0.97 & 96.29±0.72 & 74.79±1.18 & 69.13±2.28 \\
A-ConvNet-CFAR & 94.91±0.39 & 93.55±0.95 & 62.60+2.63 & 89.61±0.44 & 93.76±0.41 & 91.41±1.25 & 93.44±0.61 & 80.54±0.62 & 85.25±1.12 \\
ECCNet & 98.73±0.14 & 94.07±0.68 & 50.19+1.83 & 73.32±0.74 & 90.96±1.57 & 97.29±0.42 & 93.60±0.50 & 84.02±0.04 & 76.92±1.22 \\
ECCNet-CFAR & 96.66±0.21 & 94.34±0.77 & 60.59+0.73 & 90.90±0.52 & 95.69±0.24 & 90.79±2.36 & 88.57±1.75 & 85.27±0.59 & 83.41±1.95 \\
MVGGNet & 98.34±1.20 & 91.04±3.30 & 57.74+2.26 & 71.63±3.40 & 76.42±2.58 & 96.81±0.72 & 95.05±1.09 & 79.90±2.64 & 73.50±2.57 \\
SNINet & 95.69±1.42 & 88.41±2.74 & 69.06+1.82 & 67.47±2.27 & 88.94±1.73 & 95.97±1.81 & 95.53±1.84 & 69.06±2.71 & 67.01±2.52 \\
TDDA & 98.64±0.27 & 93.05±1.35 & 54.34+4.99 & 79.85±1.52 & 96.82±0.37 & 97.66±0.22 & 97.96±0.30 & 81.64±1.06 & 73.05±1.98 \\
HDANet (ours) & \textbf{99.64±0.13} & \textbf{96.26±0.44} & \textbf{\textbf{76.75+3.46}} & \textbf{\textbf{\textbf{\textbf{92.72±0.46}}}} & \textbf{\textbf{97.38±0.86}} & \textbf{\textbf{98.49±0.51}} & \textbf{\textbf{98.36±0.22}} & \textbf{\textbf{86.51±1.61}} & \textbf{\textbf{\textbf{\textbf{94.78±0.65}}}} \\
\bottomrule
\multicolumn{10}{l}{\begin{tabular}[c]{@{}l@{}}Note: The \textbf{\textbf{bold}} number denotes the best result. All results are the mean overall accuracy (\%) ± standard deviation over 5 runs. EOC-Azimuth/Gaussian/\\Random/Occlusion results are under the most challenging parameter in our settings (see Fig. \ref{result_eoc_curve} for detailed accuracy curves under different parameters).\end{tabular}} \\
\end{tabular}
\end{table*}

\begin{figure}[tb]
\centering
\includegraphics[width=9.5cm]{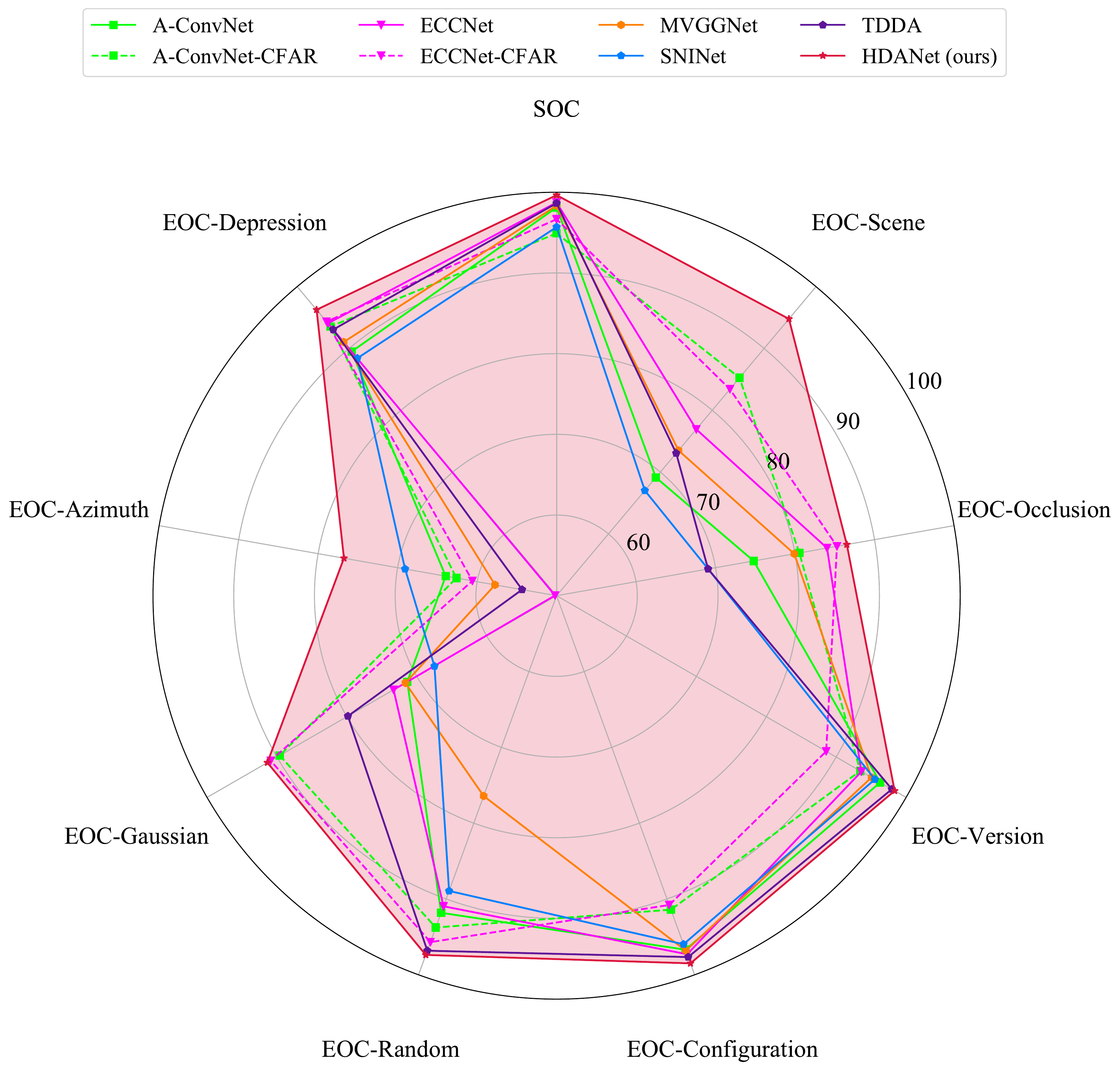}
\caption{Radar charts of experimental results (see Table \ref{table_result} for detailed numbers). Our method performs more robustly than others under various operating conditions.}
\label{result_radarmap}
\end{figure}

\begin{figure*}[tb]
\centering
\includegraphics[width=18.1cm]{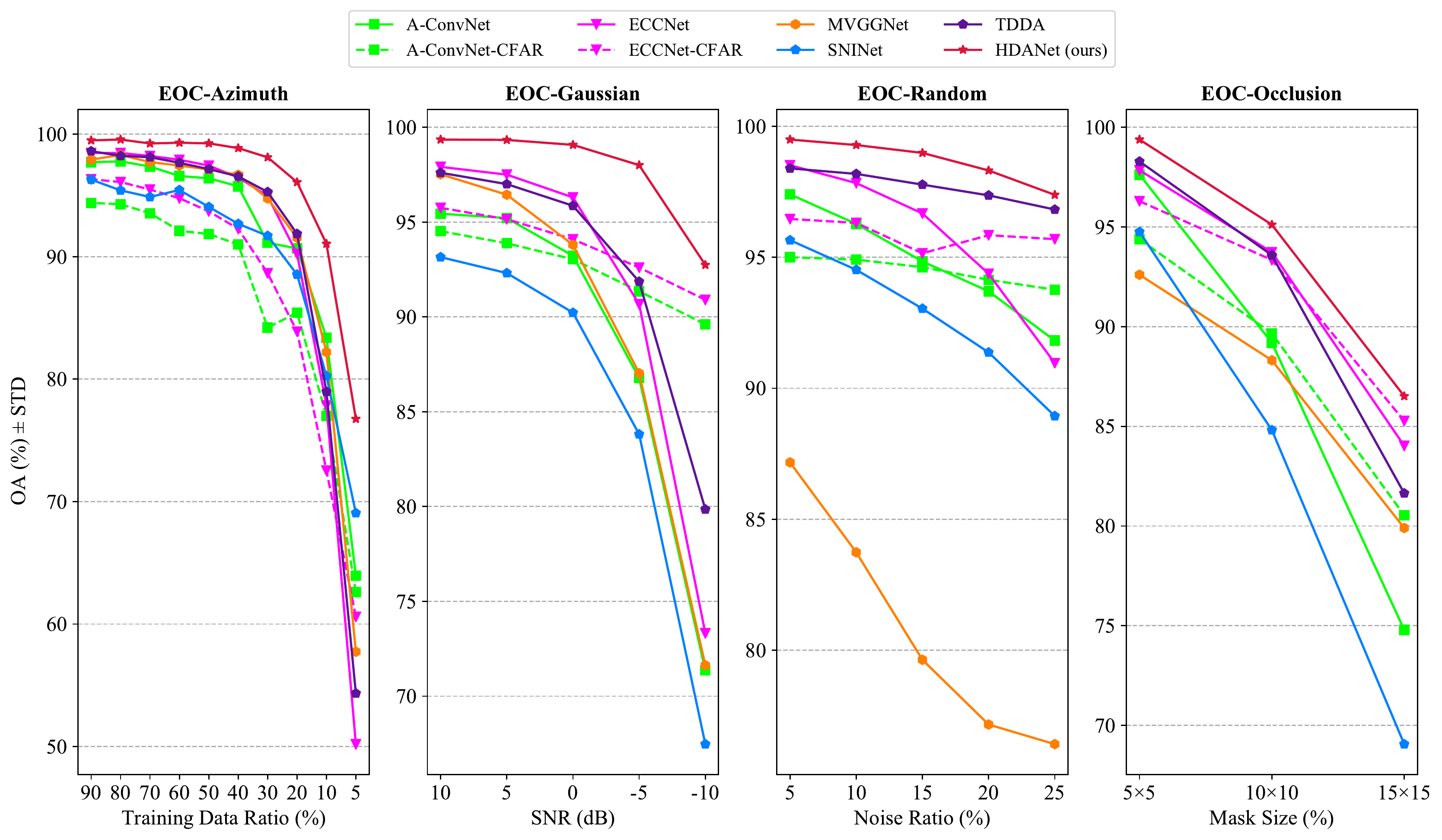}
\caption{Experimental results of different parameters under EOC-Azimuth, EOC-Gaussion, EOC-Random, and EOC-Occlusion. The accuracy decreases as the perturbation intensity increases, and our method maintains better robustness under different parameters than others.}
\label{result_eoc_curve}
\end{figure*}

In this section, we evaluated the robustness of HDANet in the nine operating conditions of MSTAR compared with other methods. The recognition results of these methods are shown in Table \ref{table_result} and Fig. \ref{result_radarmap}, and the detailed results are given as follows:

\subsubsection{\textbf{Result under SOC}} 
Because of little difference between training and test set distribution under SOC, deep learning-based approaches effectively learn correlations in the dataset, nearing saturation performance. Removing clutter and shadows from input images by CFAR causes a slight decrease in the accuracy rate (3.21\% for A-ConvNet and 2.07\% for ECCNet). The decline is due to the background correlation\cite{kechagias2021automatic, li2023discovering} in the MSTAR dataset and the shadows containing the target's structural properties\cite{ref30, li2023discovering,choi2022fusion}. Previous work\cite{ref25} pointed out that using a hard constraint mask for segmented images causes deep learning to pay attention to boundaries rather than target signatures, so a soft learnable constraint mask was developed for our deep learning-based framework. Good performance can also be achieved under SOC with transfer learning (98.34\% for MVGGNet), demonstrating the generality of deep learning's underlying feature extraction. CNN feature maps used by SNINet (95.69\%) perform lower than other methods (98.64\% for TDDA and 99.64\% for HDANet), which indicates the $l_2$ loss ignores the two-dimensional structural information in the feature maps. Therefore, we use capsule vectors to represent feature structure information with 99.64\% performance. The corresponding cosine similarity and SimSiam structure mitigate domain alignment to impair feature discrimination.

\subsubsection{\textbf{Result under EOC-Depression}} 
The increased depression angle changes the target signature and enhances the clutter, so accuracy rates improve (4.05\% for A-ConvNet and 0.27\% for ECCNet) with CFAR. Transfer learning cannot handle distribution shifts in a small dataset, achieving 91.04\% for MVGGNet. Domain alignment relies on a powerful encoder, with SNINet (88.41\%) and TDDA (93.05\%) performing below ECCNet (94.07\%). Therefore, we apply domain alignment to an encoder consisting of CNN and capsule layers to achieve a performance of 96.26\%.

\subsubsection{\textbf{Result under EOC-Azimuth}} 
Reducing the training set data loses some target signatures under partial azimuth. As shown in Fig. \ref{result_eoc_curve}, the design of a powerful encoder and domain alignment alone can not solve the robustness problem of azimuth variations. A powerful encoder (A-ConvNet, ECCNet, and MVGGNet) can overfit small data, reducing generalization performance. Domain alignment (SNINet and TDDA) may learn the same feature representation more easily with fewer data, reducing feature representation discrepancy. Data augmentation simulates local perturbations and are not effective with large variations in very few samples. However, domain alignment (TDDA, HDANet) can improve the effectiveness of data augmentation for EOC-Azimuth. HDANet strikes a balance between robustness and feature discrepancy. HDANet extracts robustness target features by domain alignment. Moreover, the SimSiam structure reduces domain alignment impairment on feature discrepancy.

\subsubsection{\textbf{Result under EOC-Gaussion}} 
Additive Gaussian noise weakens target signatures and changes the relative values of the strong and weak scattering points. Fig. \ref{result_eoc_curve} shows that the curve decreases slowly after removing clutter with CFAR because it extracts the target region blurred by noise. On the other hand, domain alignment (TDDA) is also better than other methods at low SNR. Our method combines mask and domain alignment with the best 92.72\% accuracy, using mask disentanglement to extract target features and domain alignment to enhance robustness. However, target feature extraction is difficult at low SNR (-10 dB), and data augmentation parameters need to be adjusted for the learnable target mask. It can be found that our method becomes significantly degraded below -5 dB. The CFAR method also has this problem in region extraction, but we used the masks of CFAR from the original images for simplicity under any SNR, and therefore the trend of the CFAR method is more stable below -5 dB.

\subsubsection{\textbf{Result under EOC-Random}} 
Random noise simulates a different proportion of strong clutter point interference. As shown in Fig. \ref{result_eoc_curve}, CFAR can effectively reduce clutter interference. ECCNet performs well at low proportions, but when interference increases, the attention module in ECCNet cannot effectively remove a large proportion of strong clutter. The use of fully connected layers with domain alignment (TDDA) can also reduce this interference. However, using domain alignment with CNN features (SNINet) is more likely to receive strong clutter point interference, which may be because multi-layer CNNs have aggregated information from larger receptive fields with strong clutter points from different locations. MVGGNet also has this problem. Our method suppresses clutter and extracts robust target features with the best performance.

\subsubsection{\textbf{Result under EOC-Configuration}} 
Although the T-72 configuration in the test set differs from the training set, the impact of this variant on deep learning is small at a resolution of 0.3 m, and almost all methods achieve over 95\% accuracy. However, the performances of A-ConvNet and ECCNetet decrease after extracting the target region by CFAR due to removing the T-72 gun barrel signature in the shadow. Therefore, we use a soft-constrained mask and enhance robustness with domain alignment to achieve an accuracy of 98.49\%. The learned mask can extract shadow edge information (see Fig. \ref{fig_med} for details).

\subsubsection{\textbf{Result under EOC-Version}} 
Version variants impact recognition similarly to configuration variants. Therefore, the analysis for such variants is similar to the configuration variants. However, there is a 3.69\% decrease in the accuracy of ECCNet, which may be due to the attention module in ECCNet misfitting the unstable background correlation. The scene of 4 T72 variants in the test set has a small different from the training set.

\subsubsection{\textbf{Result under EOC-Occlusion}} 
The occlusion setting affects the pixel values of the target signature and adjacent regions. As shown in Fig. \ref{result_eoc_curve}, the occlusion of sizes from 5 to 15 impacts target signatures significantly due to the small vehicle target size. Convolutional features (A-ConvNet and SNINet) are susceptible to occlusion, while the fully connected layer (MVGGNet) is less susceptible. The least is using a capsule network (ECCNet) for classification. Therefore, we use the capsule network as a classifier and design a corresponding domain alignment method, achieving 86.51\% accuracy. The paper\cite{feng2022electromagnetic} used electromagnetic scattering features to enhance the deep learning features with 82.39\% accuracy under a single look in the same setting. Our proposed method improves the robustness of deep learning to occlusion as well. However, it is challenging to achieve robust recognition algorithms with heavily obscured target signatures. We argue that using multiple-look images is a more effective way to deal with this interference since the occlusion is closely related to the relative positions of the target and the sensor. 

\subsubsection{\textbf{Result under EOC-Scene}} 
The clutter in different backgrounds significantly differs and affects the adjacent target signature. Data-driven models like deep learning have the potential to misuse the features of strong clutter. Compared to the simulated noise setting, EOC-Scene used the measured data from different scenes to further research this problem. As shown in Table \ref{table_result}, the robustness to target signature changes of A-ConvNet and ECCNet after removing clutter needs improvement. Pre-training model MVGGNet also cannot handle the data bias and distribution shifts in this downstream task. Other domain alignment methods cannot address this problem since they ignore background clutter interfering with the final features. Our method performs domain alignment of target features with mask disentanglement. Therefore, our approach achieves robustness under this new experimental setting with 94.78\% accuracy.

As discussed above, our method has significant robustness than other methods across various experimental settings in Table \ref{table_result}, Fig. \ref{result_radarmap}, and Fig. \ref{result_eoc_curve}. Compared to deep learning models (A-ConvNet and ECCNet) with CFAR or transfer learning (MVGGNet), our approach improves robustness through domain alignment of target features. We achieve feature disentanglement and design a novel domain alignment framework for SAR recognition compared with other domain alignment methods (SNINet and TDDA).

\subsection{Analysis}
\label{Analysis}
In this subsection, we conducted numerous qualitative and quantitative analyses to validate the effectiveness of our approach in robust recognition and clutter suppression. These analyses encompassed the ablation study, qualitative research, and the analysis of the clutter effect.

\subsubsection{\textbf{Ablation study}}
\begin{figure}[!tb]
\centering
\includegraphics[width=8.8cm]{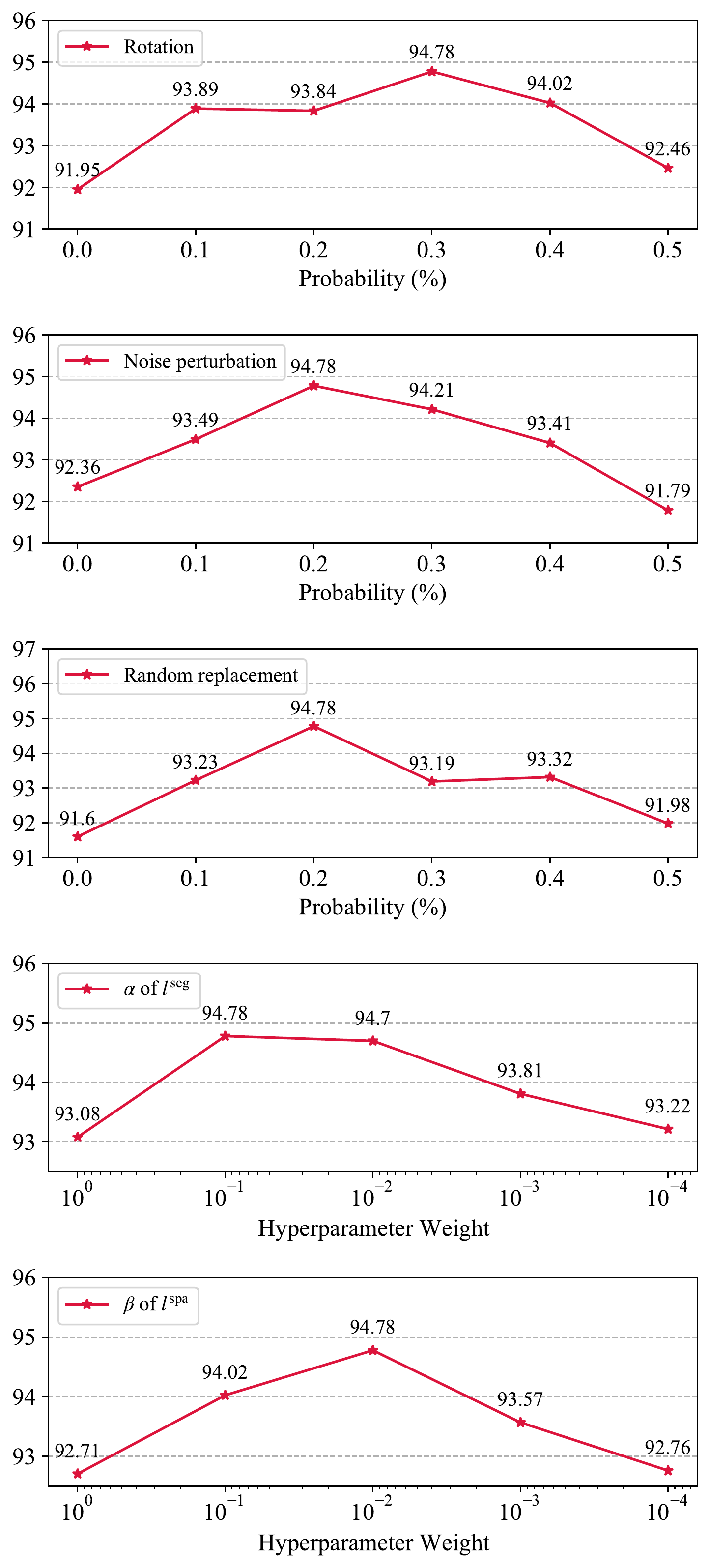}
\caption{Impacts of the hyperparameters under EOC-Scene. From top to bottom are the probabilities of the three data augmentation methods (rotation, noise perturbation, and random replacement) in domain generation and the hyperparameters of the two auxiliary tasks (segmentation task and sparse loss) in mask disentanglement. These methods are not highly sensitive to hyperparameters and are effective in relatively large intervals.}
\label{result_hyperparameter}
\end{figure}

Since the target signature and the background clutter change significantly under EOC-scene, we performed the ablation study and hyperparameter discussion under EOC-Scene in Table \ref{tab_ablation} and Fig. \ref{result_hyperparameter}. 

\begin{table}[!tb]
\centering
\caption{Ablation Study under EOC-Scene. The baseline is a CNN Encoder and a Capsule Classifier. DDG is Domain Data Generation. MMD is Multitask-assisted Mask Disentanglement. DATF is Domain Alignment of Target Features}
\label{tab_ablation}
\renewcommand\arraystretch{1.25}
\begin{tabular}{ccccc} 
\toprule
Baseline & \multicolumn{1}{c}{DDG} & MMD & DATF & OA (\%) ± STD~$\uparrow$ \\ 
\hline
$ \checkmark$ & \multicolumn{1}{c}{} & ~ & ~ & 86.61±7.31 \\
$ \checkmark$ & $\checkmark$ & & ~ & 87.13±4.10 \\
$ \checkmark$ & $\checkmark$ & $\checkmark$ & & 92.29±2.46 \\
$ \checkmark$ & $\checkmark$ & $ \checkmark$ & $ \checkmark$ & \textbf{\textbf{94.78±0.65}} \\
\bottomrule
\multicolumn{5}{l}{\begin{tabular}[c]{@{}l@{}}Note: The \textbf{\textbf{bold}} number denotes the best result. All \\results are the mean overall accuracy (\%) ± standard\\ deviation over 5 runs.\end{tabular}} \\
\end{tabular}
\end{table}

\textbf{Baseline} used a modified CNN encoder of U-Net and a capsule classifier, and its performance is 86.61\%. Compared with ECCNet, we removed the attention module and image reconstruction task to avoid enhancing the overfitting for background clutter. The baseline improvement illustrates the need for successful modules in natural images to be refined for challenges in SAR images. However, the large standard deviation (7.31\%) indicates that deep learning still overfits the background clutter. 

\textbf{Effects of domain data generation.}
With data augmentation on the baseline in Table \ref{tab_ablation}, the accuracy is improved to 87.13\% with a smaller standard deviation (4.10\%). Then, we discussed the detailed effect of different data augmentation methods with other modules under EOC-Scene in Fig \ref{result_hyperparameter}. Specifically, different data augmentation methods improve domain alignment, but the high probability rate can cause alignment centers to deviate from the original image, which reduces recognition accuracy. We advise setting each probability to [0.1, 0.4] and the total probability close to 0.5 for domain alignment. To the best of our knowledge, at least ten data augmentation methods have been proposed for SAR vehicle target recognition, and our method aims to simulate local variations of target signatures for domain alignment. Although we validated the effectiveness of the proposed method under various experimental settings, it is still worth exploring how to use data augmentation and generation more effectively for complex operating conditions and small datasets.

\textbf{Effects of multitask-assisted mask disentanglement.} 
We used the multitask setting to extract the feature layer mask, which further suppresses clutter and achieves an accuracy of 92.29\%±2.46 in Table \ref{tab_ablation}. The learnable masks can be better integrated with deep learning and exploit the good properties of the middle layer\cite{ref44}, which avoids strong clutter points in the input image. Moreover, the multitask setting introduces the target location and sparse priors to solve the background correlation in a small dataset. As shown in Fig \ref{result_hyperparameter}, we discussed the hyperparameter setting of mask disentanglement under EOC-Scene. Specifically, $\alpha$ and $\beta$ control the impact of the segmentation task and sparse loss. We advise setting $\alpha$ and $\beta$ to [1e-1, 1e-3]. Hyperparameters below these ranges reduce the auxiliary task effect, and above these ranges impair the primary recognition task and discrimination of the target features. The background interference in small SAR vehicle datasets is more severe compared to large datasets in computer vision because SAR images are collected under specific operating conditions. Therefore, this module achieves feature disentanglement to ensure correct feature representation. 

\textbf{Effects of domain alignment of target features.}
Considering the connection between causality and invariance\cite{ref67}, HDANet uses domain alignment of target features further enhance the robustness of features. Its performance is impressive, reaching 94.78\% with the smallest standard deviation (0.65\%). In the domain alignment module, we used capsule vectors to preserve feature space information and cosine similarity as the contrastive loss. The SimSiam structure increases inter-class distance and mitigates the conflict between contrastive loss and classification loss. By realizing feature disentanglement and alignment through the above three modules, we established a new domain alignment framework for robust SAR target recognition, which ensures the causality and robustness of feature representations.

\subsubsection{\textbf{Qualitative research}} 
We illustrated the effectiveness of our proposed method with visualization results and methods, including feature separability, intermediate results, and segmentation results.

\textbf{Feature separability.} 
\begin{figure}[!tb]
\centering
\includegraphics[width=8.8cm]{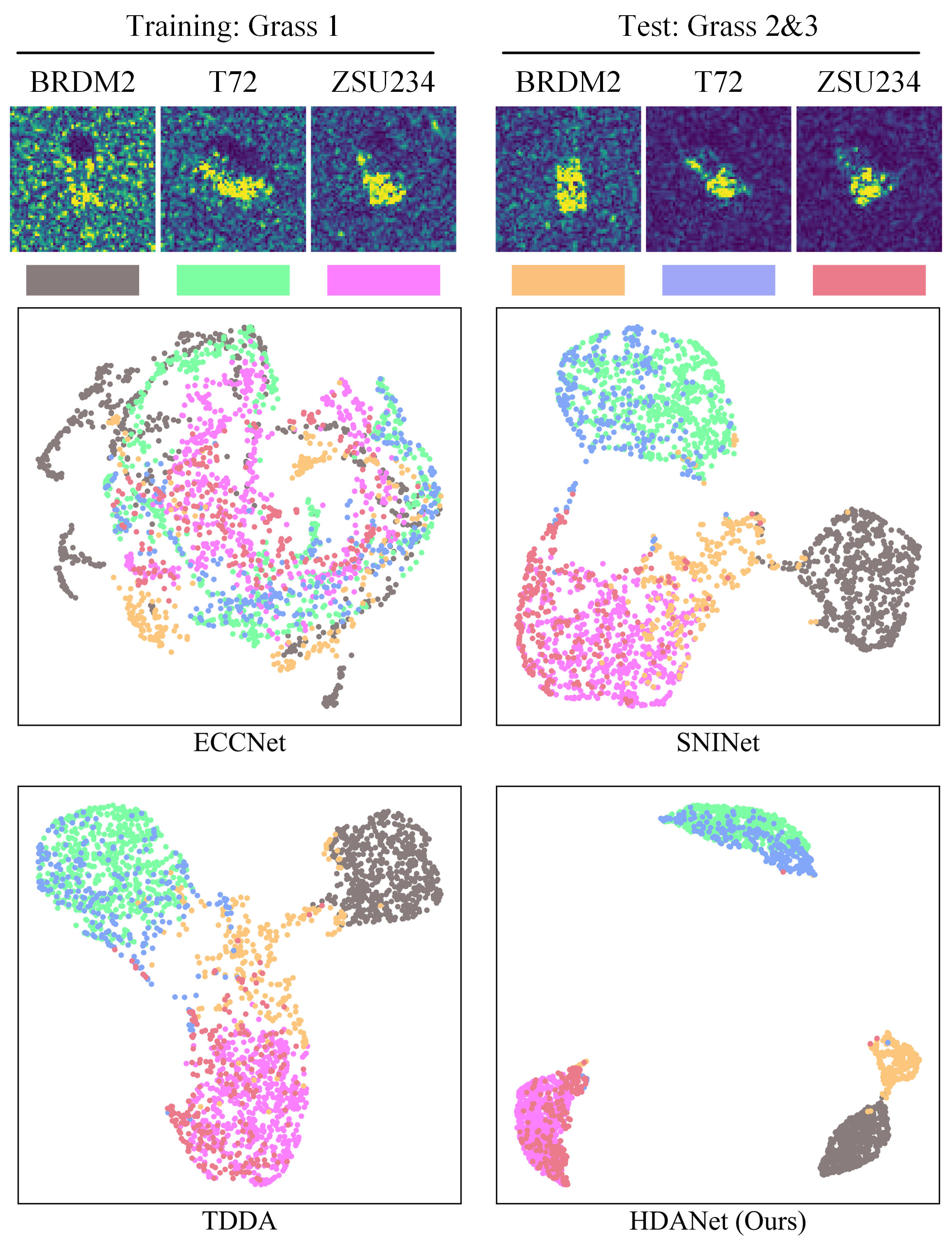}
\caption{Visualization of feature separability across models under EOC-Scene. We used UMAP\cite{mcinnes2018umap} to visualize the penultimate layer features of each model. We can see that the domain alignment (SNINet and TDDA) method has a good clustering effect than ECCNet in the training set, and our method further improves the robustness of domain alignment methods to distribution shifts in SAR using feature disentanglement.}
\label{result_umap}
\end{figure}
Using the EOC-Scene as an example, we visualized the features of the penultimate layer of different models by uniform manifold approximation and projection (UMAP)\cite{mcinnes2018umap}. From Fig. \ref{result_umap}, we can see that the domain alignment (SNINet and TDDA) methods produce good clustering results than ECCNet, resulting in larger inter-class distances for improved robustness. However, the features extracted by the domain alignment methods contain unstable clutter due to background interference in the SAR images, which reduces the robustness of these methods to complex distribution shifts in SAR. As a result, these domain alignment methods yielded large interclass distances, but incorrect feature representations resulted in lower accuracy than ECCNet. Our method extracts target features by mask disentanglement, suppressing unstable clutter from interfering with the final features and improving the robustness using domain alignment of target features. Despite the success of our framework, the feature shift in BRDM2 visualization results shows that there are still opportunities for further enhancements.

\begin{figure}[!tb]
\centering
\includegraphics[width=8.8cm]{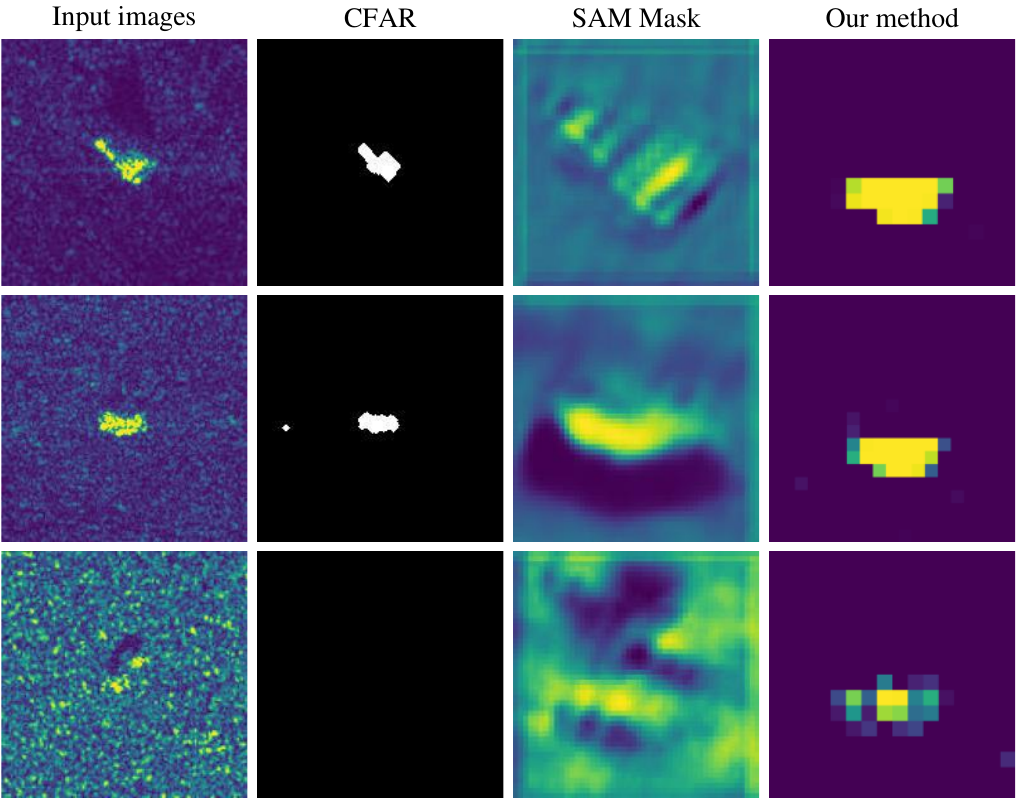}
\caption{Intermediate masks of different methods. We can observe that the feature layer mask avoids strong clutter interference in the input image, while our method solves the problem of spatial attention mechanism overfitting background clutter.}
\label{fig_masks}
\end{figure}
\textbf{Intermediate masks of different methods} in Fig. \ref{fig_masks} show that our methods solved the problem of background overfitting caused by data bias. Threshold methods, such as CFAR, are susceptible to strong clutter interference in the input image, failing to detect the target in the strong clutter region. Furthermore, existing threshold methods~\cite{ref6,choi2022fusion} combine morphological operations to eliminate strong clutter points but blur fine target contours. The attention mechanism extracts the target mask at the feature layer to avoid the interference of strong clutter points. However, the spatial attention mechanism masks do not effectively suppress the background clutter due to data bias and activation function. In contrast, our method suppresses the background correlation by adding priori constraints and improves the computation of the target mask.

\begin{figure}[!tb]
\centering
\includegraphics[width=8.8cm]{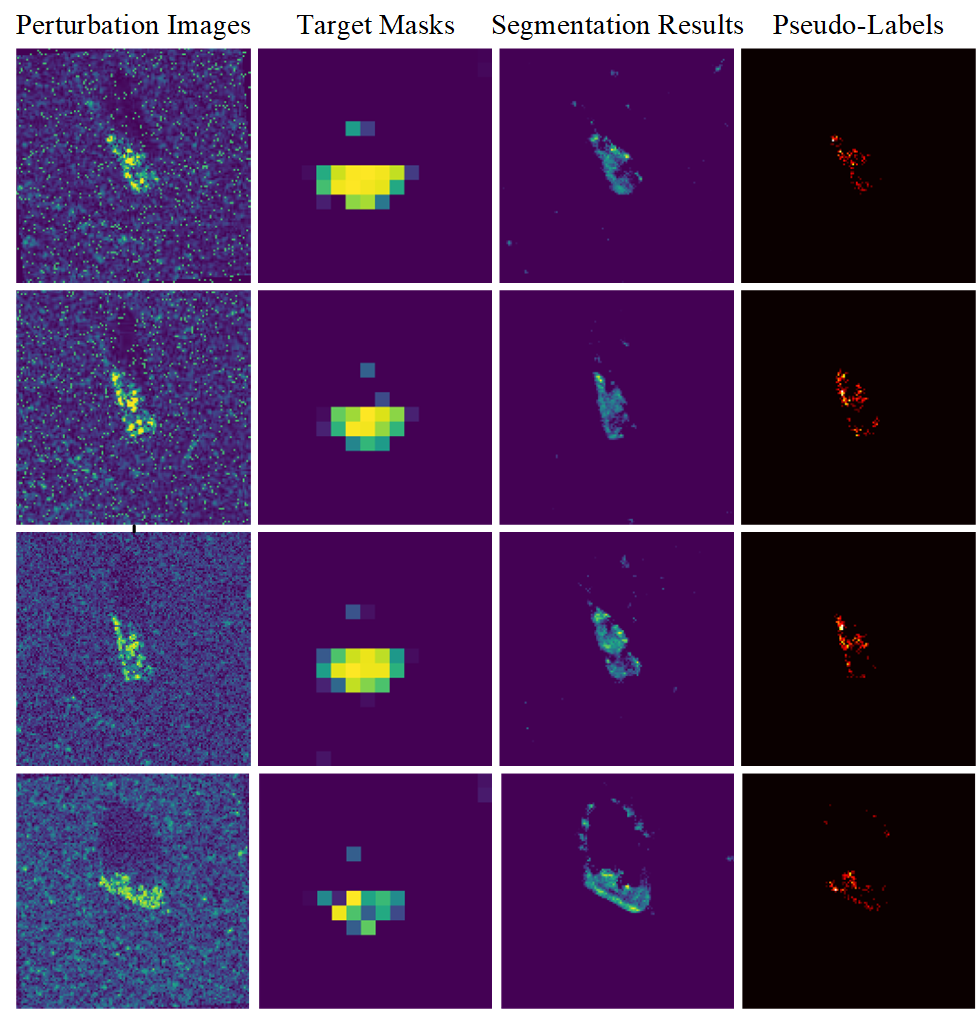}
\caption{Intermediate results visualization. We can observe that the target masks of the feature layer in the second column ignore the background clutter that occupies most of the images. And the segmentation results in the third column also separate the target from the clutter. Our results and pseudo-labels retain some shadow edges that reflect the target signatures.}
\label{fig_med}
\end{figure}
\textbf{Intermediate results} in Fig. \ref{fig_med} show that our target mask in the feature layer separates the target region from the background, and the edge information of the target shadow is contained above the target region in the mask. Our method suppresses clutter better than CBAM's masks in Fig. \ref{fig_atten}. The segmentation results also illustrate the effectiveness of our method in clutter suppression. Our way of generating pseudo-labels balances accuracy and efficiency. The weight of random clutter is reduced in the saliency maps by two averaging operations of multiple class saliency maps and SmoothGrad, and our pseudo-labels extract the valuable targets and shadow regions for recognition. And since the target mask is in the middle feature layer, the automatically generated coarse labels are sufficient for the segmentation task to play an auxiliary role with $l_1$ loss, allowing the mask to distinguish between background and target discrepancies and obtain the correct feature representation. 

\begin{figure}[!tb]
\centering
\includegraphics[width=8.8cm]{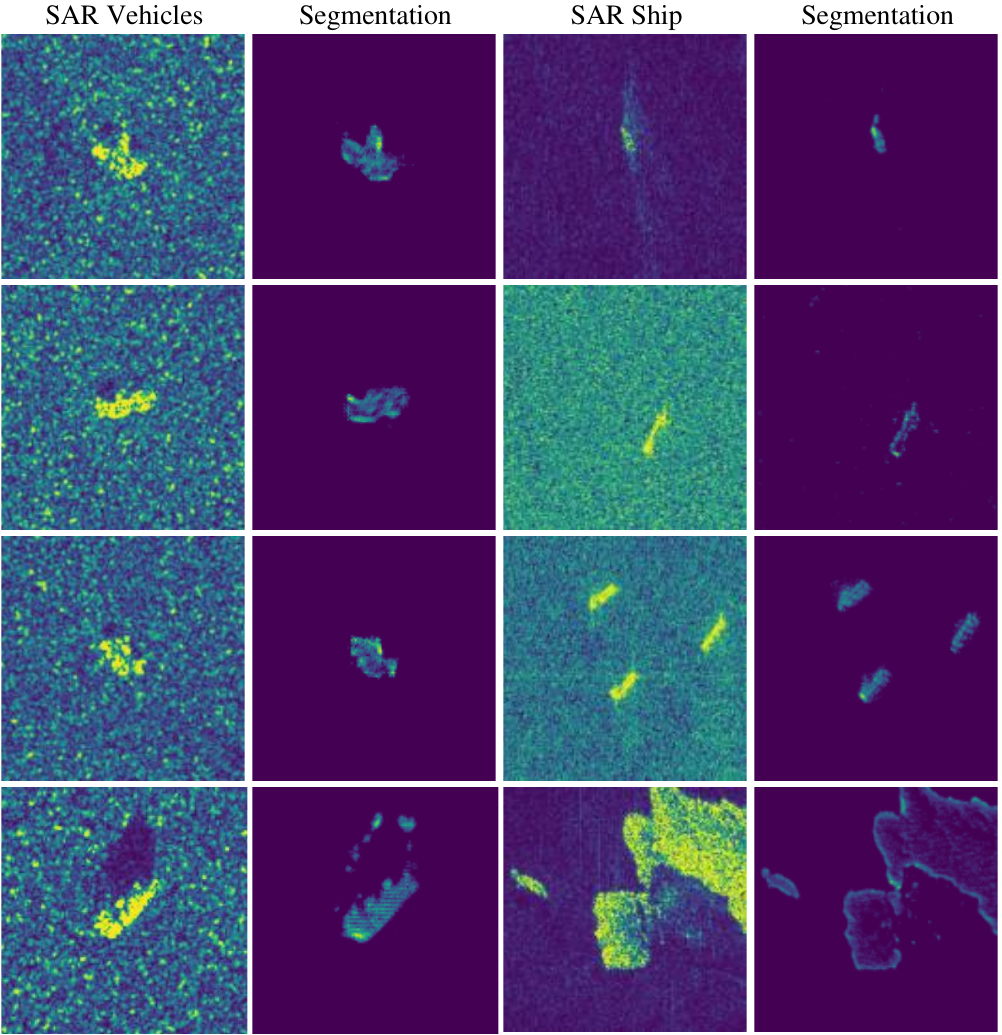}
\caption{Segmentation results (from left to right are SAR vehicle images in MSTAR\cite{ref4}, ground segmentation results, SAR ship images under complex backgrounds\cite{ref54}, and sea segmentation results). HDANet distinguishes targets from different ground/sea clutter.}
\label{result_seg_groundsea}
\end{figure}
\textbf{Segmentation results on different datasets.}
The segmentation task was used as an auxiliary task to improve the robustness of the recognition task. Therefore, we used pseudo-labels and did not pursue precise segmentation results, which balance accuracy and efficiency. Although not as accurate as manual annotation, the rough labels allow the model to learn the difference between the target and the background. In Fig. \ref{result_seg_groundsea}, we trained an HDANet on the MSTAR dataset and visualized the segmentation results of vehicles and ships\cite{ref54}. Our method identifies target regions in ground/sea clutter, demonstrating its segmentation ability to generalize to different environments. Although the MSTAR dataset targets are all centrally placed, HDANet distinguishes between targets and clutter in multi-target SAR ship slices with 256 × 256 pixels. However, the pseudo-label with the saliency map makes HDANet focus on the structural edges of targets, similar to edge detection\cite{ref56}, causing large scene edges preserved in Fig. \ref{result_seg_groundsea}.

\subsubsection{\textbf{Analysis of the clutter effect on deep learning}}
Clutter suppression is important for robust SAR vehicle recognition due to blurry target signatures. Therefore, it is necessary to analyze whether deep learning uses the correct features for recognition, \emph{i.e.}, using only the target signatures rather than clutter correlations. However, it is difficult to capture the negative impact of overfitting clutter on accuracy in experiments with similar backgrounds. Consequently, we used Saliency maps and the Shapley value below to qualitatively and quantitatively analyze the degree of deep learning overfitting background clutter, \emph{i.e.}, the feature causality.

\begin{figure*}[!tb]
\centering
\includegraphics[]{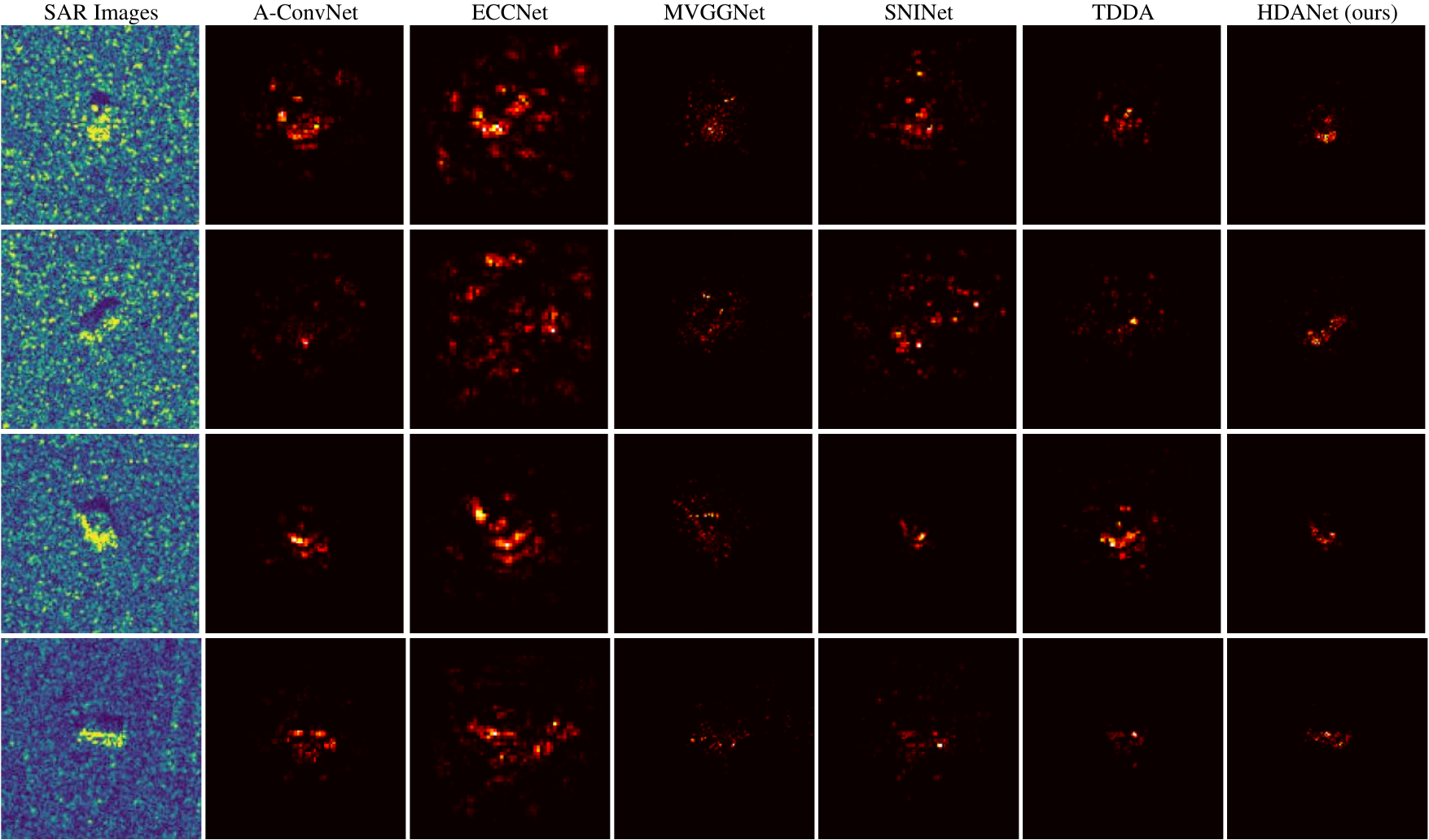}
\caption{Saliency maps of each model (from left to right are the SAR images and saliency maps of A-ConvNet, ECCNet, MVGGNet, SNINet, TDDA, and HDANet). The clutter intensity of the input image decreases in order from the top to the bottom. Saliency maps reflect the dependency of a model for each pixel point during recognition. From a visual perspective, our method focuses more on the target region, and others may use clutter.}
\label{result_saliency_map}
\end{figure*}
\textbf{Saliency maps.} We first qualitatively analyzed causality through the saliency map\cite{kokhlikyan2020captum}. The results of different up-sampling saliency maps are shown in Fig. \ref{result_saliency_map}. We can find that the models (A-ConvNet, ECCNet, and SNINet) using images with different input sizes overfit the background clutter around the target, so the center cropping does not solve the problem of clutter interference. ECCNet, with image reconstruction and attention module, focuses on clutter. Other domain alignment methods also contain some clutter regions. These results indicate that the methods successfully applied in natural images need to be improved in SAR images. Due to our novelty improvement for mask and domain alignment methods in SAR, our method focuses more on the target region than background clutter. Since the saliency map calculates the importance of a single image at the pixel level, it only responds qualitatively to causality and lacks quantitative statistical metrics of the whole region. We next used the Shapley value as the quantitative metric of the clutter effect.

\textbf{Shapley value} is the unique solution satisfying the four properties and calculates the contribution to the cooperative game fairly\cite{sundararajan2020many}. According to our previous research\cite{li2023discovering}, we segment the target and clutter regions in the input image based on CFAR as two players $\{0,1\}$ and estimate their contribution to recognition based on the Shapley value: 
\begin{equation}\label{eq_Shapley}
{\rm Sh}_i=\sum_{S\subseteq I\backslash \{i\}} \frac{\left| S \right|! \cdot (\left| I \right|-\left| S \right|-1)!}{\left| I \right|!}\left[f(S\cup\{i\})-f(S)\right],
\end{equation}
where ${\rm Sh}_i$ is the Shapley value of $i$-th player, $I$ is a set of all players, $\left| \cdot \right|$ is the number of elements in the set, $f(\cdot)$ is classification score corresponding to the true class before softmax. The baseline values of the inputs are set to 0. We average the Shapley values for all the images and calculate the proportions of clutter Shapley values for different models to roughly estimate the degree of the overfitting for clutter.
\begin{table}[!tb]
\centering
\caption{Clutter Contribution Ratio of Different Models under EOC-Scene. This Metric Reflects The Contribution of Clutter to Correct Recognition Results in The Training Set. A Lower Value Indicates That A Model is Less Dependent on Clutter for Recognition}
\label{tab_shapely}
\renewcommand\arraystretch{1.25}
\begin{tabular}{cccc} \toprule
Model & Avg. (\%) ± STD $\downarrow$ & Input size & Params \\ \midrule
A-ConvNet & 27.69±2.80 & 88 × 88 & 0.30 M \\
ECCNet & 46.71±2.77 & 64 × 64 & 7.99 M \\
MVGGNet & 55.71±3.68 & 128 × 128 & 16.81 M \\
SNINet & 25.99±5.69 & 88 × 88 & 0.35 M \\
TDDA & 34.45±1.44 & 128 × 128 & 0.80 M \\
Our method & \textbf{15.41±2.41} & 128 × 128 & 12.28 M \\ \bottomrule
\multicolumn{4}{l}{\begin{tabular}[c]{@{}l@{}}Note: The \textbf{\textbf{bold}} number denotes the best result. All results are \\the mean and standard deviation of clutter contribution ratio \\over 5 runs. Although our method uses the original image size, \\the effect of background clutter is minimal.\end{tabular}} 
\end{tabular}
\end{table}

As shown in Table \ref{tab_shapely}, due to the texture bias of CNN\cite{geirhos2018imagenet}, the texture of the clutter region is exploited by models with varying degrees. Nevertheless, our method has the least clutter effect (15.41\%) due to feature disentanglement and alignment. Comparing the different image sizes, center cropping can attenuate the effect of clutter but cannot solve this overfitting. Although ECCNet uses the smallest image size (64 × 64), image reconstruction significantly increases clutter impact (46.71\%), and its attention module doesn't solve the problem. Clutter has the greatest effect on MVGGNet (55.71\%)\footnote{Pre-training helps MVGGNet, and clutter contribution without pre-training is 65.35\%.}. This number also shows that the complex model (16.81 M) is more likely to overfit clutter than A-ConvNet (0.30 M and 27.69\%). Other domain alignment methods also cannot overcome overfitting clutter (25.99\% for SNINet and 34.45\% for TDDA), resulting in domain alignment instability with clutter interference. Although the Shapley value is a rough estimate for the effect of clutter on recognition, our analysis is sufficient to show that domain alignment in SAR needs to disentangle targets and clutter features.

\subsection{Limitations}
\label{Limitations}
While our method demonstrates satisfactory performance across a wide range of operating conditions, it is important to acknowledge its limitations. In this subsection, we present the limitations along with proposed solutions. Our method refers to the traditional detection and recognition process, treating the target and background clutter as separate entities solved by two different modules. While this separation reduces task difficulty, further integrating these two tasks has the potential to improve robustness and eliminate the current reliance on pseudo-labels and hyperparameters. Additionally, our approach heavily relies on data augmentation methods to expand the diversity of the single-domain dataset. However, the insufficient data still poses a constraint on our approach. To address this limitation, considering the increasing number of data from different SAR sensors, we plan to leverage a self-supervised learning approach to effectively extract features from a large volume of real-world SAR data. This way would enable us to mitigate the issues related to data bias and distribution shifts in small datasets.

\section{Conclusion}

This paper proposes a novel domain alignment framework named HDANet for robust SAR vehicle recognition. Its primary objective is to achieve robust recognition with feature disentanglement and alignment, and the framework comprises three essential modules: domain data generation, multitask-assisted mask disentanglement, and domain alignment of target features. Extensive experimental results conducted on the MSTAR dataset substantiate the effectiveness of the proposed approach in achieving robust recognition. The advantages and limitations of this approach are thoroughly discussed through comprehensive quantitative and qualitative analyses. Furthermore, future research directions involve investigating self-supervised learning with large amounts of SAR data from open sources to extract powerful features and address downstream task problems.

\bibliographystyle{IEEEtran}
\bibliography{ref}

\newpage
\begin{IEEEbiography}[{\includegraphics[width=1in,height=1.25in,clip]{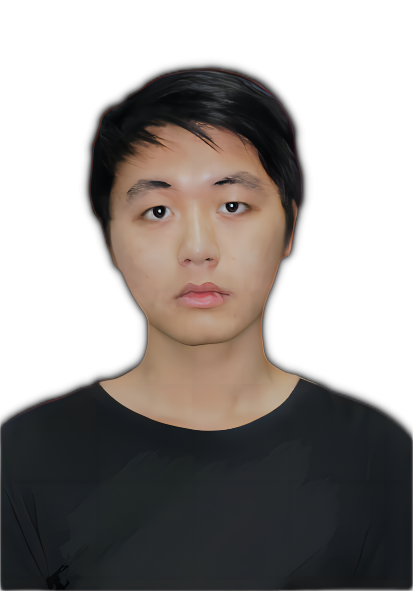}}]{Weijie LI} was born in 1998. He received the university B.E. degree in information engineering from Xi’an Jiaotong University, Xi’an, China, in 2019. He is pursuing his Ph.D. degree at the National University of Defense Technology, Changsha, China. He has published papers in respected journals, including IEEE Geoscience and Remote Sensing Letters, SCIENTIA SINICA Informationis, and Journal of Radars.
His research interests mainly focused on radar target recognition and deep learning. 
\end{IEEEbiography}
\vspace{-10mm}
\begin{IEEEbiography}[{\includegraphics[width=1in,height=1.25in,clip]{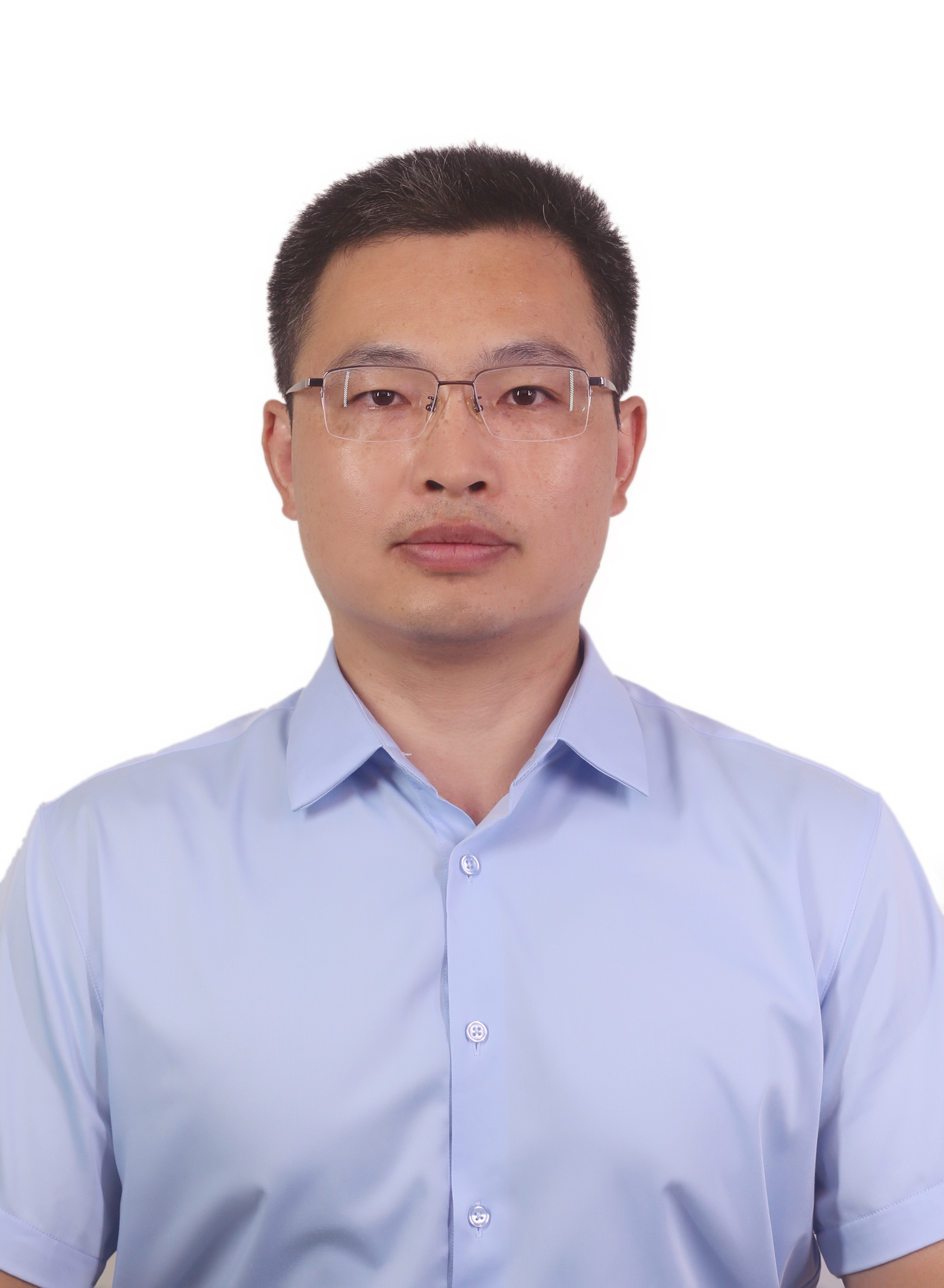}}]{Wei YANG} was born in 1985. He received the university B.E. degree from Wuhan University, Wuhan, China, in 2006, and his M.S. and Ph.D. degrees from the National University of Defense Technology in 2008 and 2012, respectively. Currently, He is an associate professor at the National University of Defense Technology. He has published numerous papers in respected journals, including IEEE Transactions on Aerospace and Electronic Systems, IET Signal Processing, and Journal of Systems Engineering and Electronics. His main research interests include cognitive radar target detection and recognition and radar target tracking.
\end{IEEEbiography}
\vspace{-10mm}
\begin{IEEEbiography}[{\includegraphics[width=1in,height=1.25in,clip]{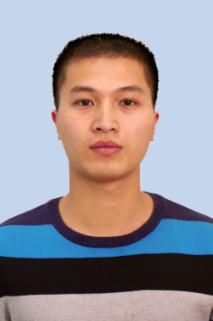}}]{Wenpeng Zhang} was born in 1989. He received the university B.E., M.S., and Ph.D. degrees from the National University of Defense Technology in 2012, 2014, and 2018, respectively. Currently, He is an assistant professor at the National University of Defense Technology. He has published numerous papers in respected journals, including IEEE Transactions on Aerospace and Electronic Systems, IEEE Signal Processing Letters, and IEEE Geoscience and Remote Sensing Letters.
His research interests include radar signal processing, parameter estimation, and sparse signal representation.
\end{IEEEbiography}
\vspace{-10mm}
\begin{IEEEbiography}[{\includegraphics[width=1in,height=1.25in,clip]{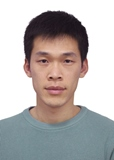}}]{Tianpeng Liu} received the B.E. and M.E and Ph.D. degrees from the National University of Defense Technology, Changsha, China, in 2008, 2011, and 2016 respectively. He is currently an associate professor at the College of Electronic Science and Technology. He has published numerous papers in respected journals, including IEEE Transactions on Aerospace and Electronic Systems and International Conference on Signal Processing. His primary research interests are radar signal processing, electronic countermeasure, and cross-eye jamming.
\end{IEEEbiography}
\vspace{-10mm}
\begin{IEEEbiography}[{\includegraphics[width=1in,height=1.25in,clip]{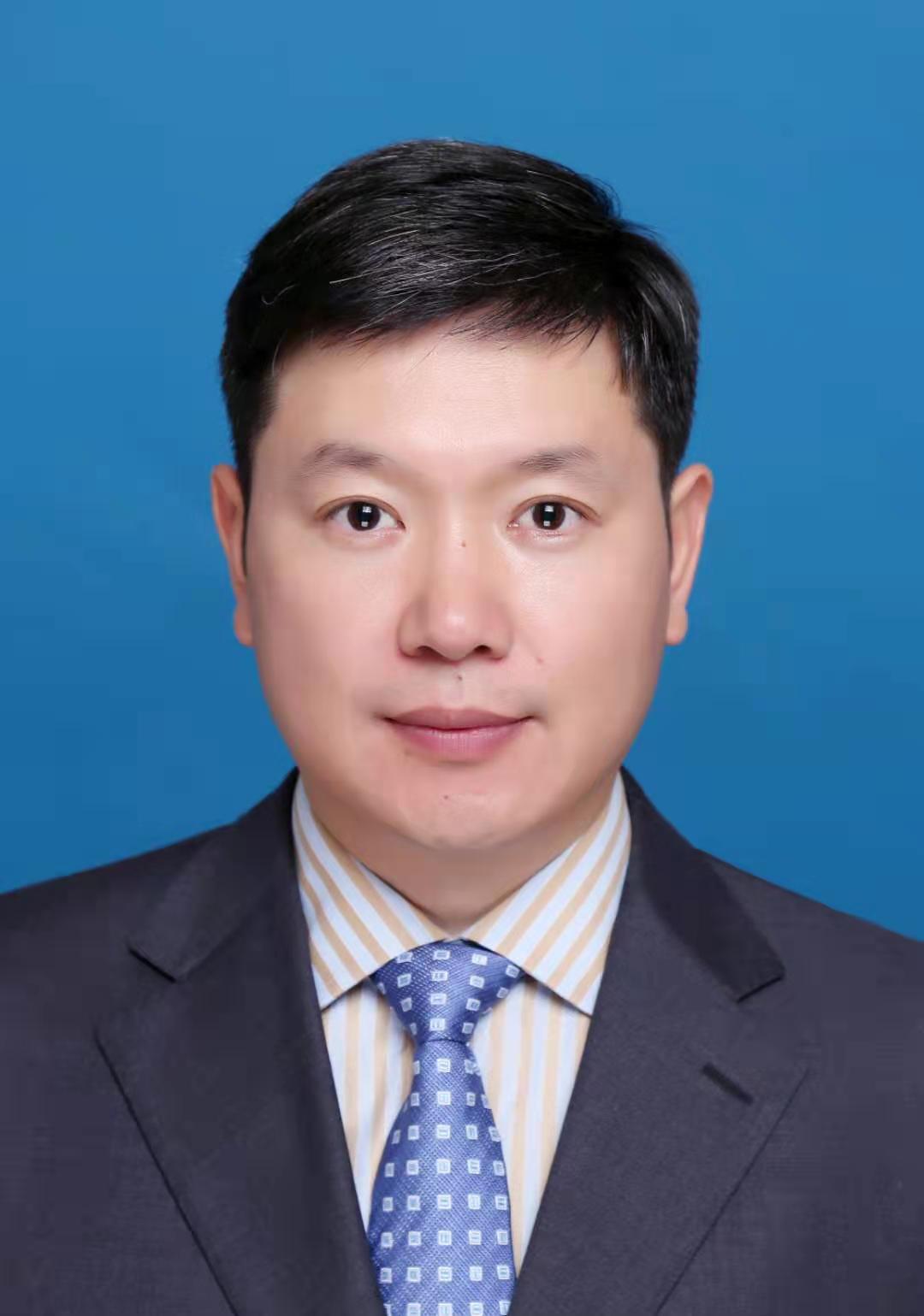}}]{Yongxiang LIU} received the B.S. and Ph.D. degrees from the College of Electronic Science at the National University of Defense Technology, Changsha, China, in 1997 and 2004, respectively. He is currently a Full Professor at the National University of Defense Technology. He has published numerous papers in respected journals, including IEEE Transactions on Image Processing, IEEE Transactions on Geoscience, and Remote Sensing. His research interests mainly include radar imaging, SAR image interpretation, and artificial intelligence. He is a member of the IEEE.
\end{IEEEbiography}
\vspace{-10mm}
\begin{IEEEbiography}[{\includegraphics[width=1in,height=1.25in,clip]{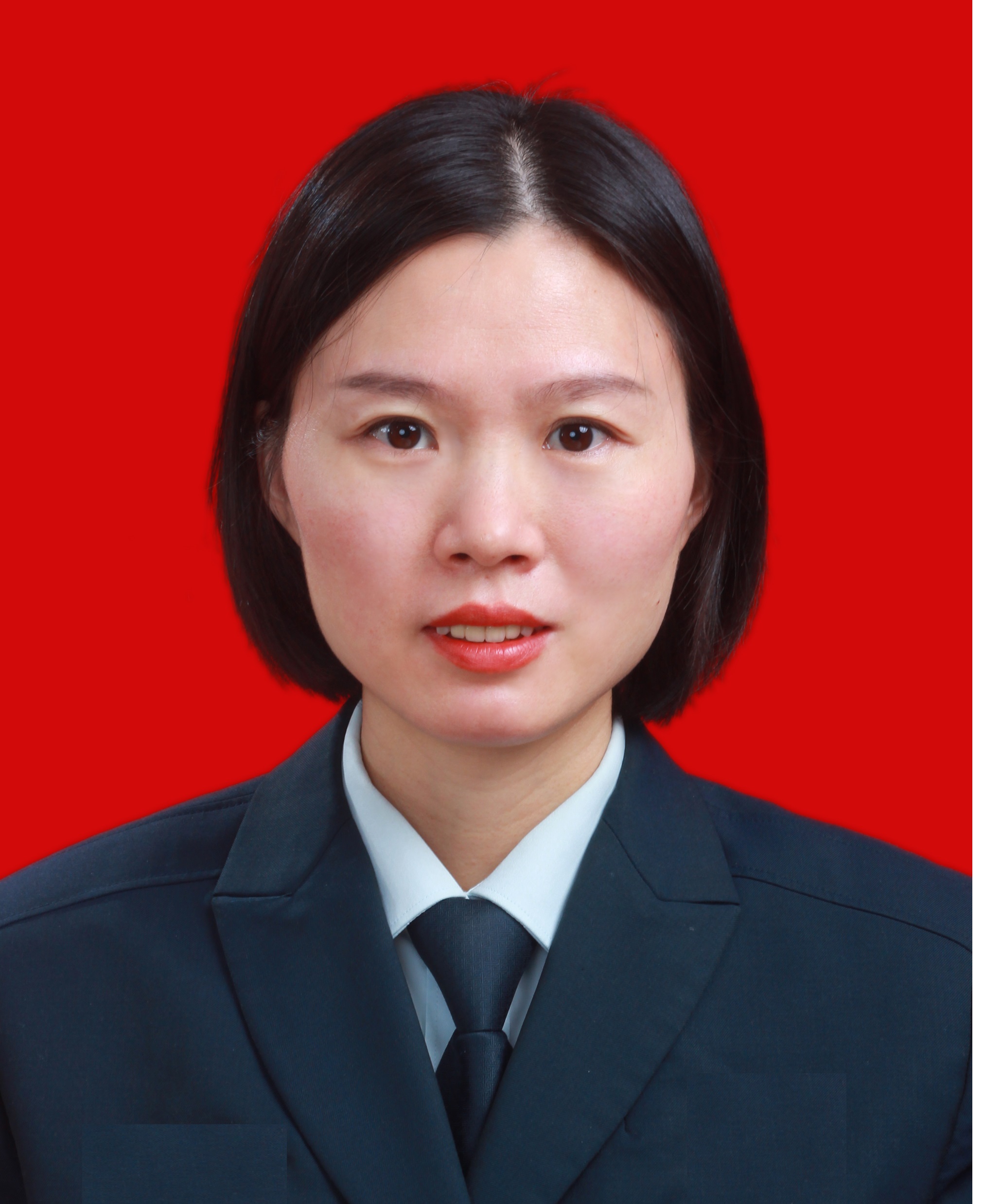}}]{Li Liu} (Senior Member, IEEE) received her Ph.D. degree from the National University of Defense Technology (NUDT), China, in 2012. 
She is now a Full Professor with NUDT. She has held visiting appointments at the University of Waterloo in Canada, at the Chinese University of Hong Kong, and at the University of Oulu in Finland. Dr. Liu served as a co-chair of many International Workshops along with major venues like CVPR and ICCV. She served as the leading guest editor of the special issues for IEEE TPAMI and IJCV. 
She also served as an Area Chair for several respected international conferences. She currently serves as an Associate Editor for IEEE TCSVT and Pattern Recognition. Her research interests include computer vision, pattern recognition, and machine learning. Her papers currently have over 10,000 citations, according to Google Scholar.
\end{IEEEbiography}

\end{document}


\title{Hierarchical Disentanglement-Alignment Network for Robust SAR Vehicle Recognition:

Supplementary material}
\author{Weijie Li, Wei Yang, Wenpeng Zhang, Tianpeng Liu, Yongxiang Liu, Li Liu
\thanks{The authors are with the College of Electronic Science and Technology, National University of Defense Technology, Changsha, 410073, China (email: lwj2150508321@sina.com, yw850716@sina.com, zhangwenpeng08@nudt.edu.cn, everliutianpeng@sina.cn, lyx\_bible@sina.com, and dreamliu2010@gmail.com).}}

\markboth{This manuscript was submitted to IEEE Journal of Selected Topics in Applied Earth Observations and Remote Sensing}%
{Li \MakeLowercase{\textit{et al.}}: A Sample Article Using IEEEtran.cls for IEEE Journals}
\maketitle

\section{A brief statistical table of the MSTAR} 

The MSTAR folders are named depression angle (DEG), collection (COL 1/2), scene (SCENE 1/2/3), target, and variants. As mentioned in\cite{ref58}, the target data includes three different locations (New Mexico, northern Florida, and northern Alabama), one of which (northern Florida) has two collection points at different times of the year (May and November). All target sites are grass-covered and flat.
The data file is divided into two parts, the first is about the annotation (ground truth) and sensor information, and the second is about the magnitude block and the phase block of SAR images.

MSTAR-Predict Lite Software can simulate floating point greyscale SAR target signatures in background clutter and noise. The software allows depression angle and rotation angle variations for the following four public targets (T72, BMP2, BTR70) and reference target (SLICY).

MSTAR-Public Clutter contains clutter file images, 100 files at a 15° depression angle.

MSTAR-Public Mixed Targets contain seven public targets (2S1, BRDM2, BTR60, D7, T62, ZIL131, ZSU234) and reference target (SLICY) in Table \ref{mstar1}.

\begin{table*}[!htb]
\caption{Number of MSTAR-Public Mixed Targets}
\label{mstar1}
\centering
\renewcommand\arraystretch{1.3}
\begin{tabular}{|c|c|c|c|c|c|c|c|c|c|c|} \hline
 & COL1 & \multicolumn{9}{c|}{COL2} \\ \cline{2-11}
DEG & SCENE1 & \multicolumn{7}{c|}{SCENE1} & SCENE2 & SCENE3 \\ \cline{2-11}
 & BTR60 & 2S1 & BRDM2 & D7 & T62 & ZIL131 & ZSU234 & SLICY & ZSU234 & BRDM2 \\ \hline
15° & 195 & 274 & 274 & 274 & 273 & 274 & 274 & 274 & ~ & ~ \\ \hline
16° & ~ & ~ & ~ & ~ & ~ & ~ & ~ & 286 & ~ & ~ \\ \hline
17° & 256 & 299 & 298 & 299 & 299 & 299 & 299 & 298 & ~ & ~ \\ \hline
29° & ~ & ~ & ~ & ~ & ~ & ~ & ~ & 210 & ~ & ~ \\ \hline
30° & ~ & 288 & 287 & ~ & ~ & ~ & 288 & 288 & 118 & 133 \\ \hline
31° & ~ & ~ & ~ & ~ & ~ & ~ & ~ & 313 & ~ & ~ \\ \hline
43° & ~ & ~ & ~ & ~ & ~ & ~ & ~ & 255 & ~ & ~ \\ \hline
44° & ~ & ~ & ~ & ~ & ~ & ~ & ~ & 312 & ~ & ~ \\ \hline
45° & ~ & 303 & 303 & ~ & ~ & ~ & 303 & 303 & 119 & 120 \\ \hline
\end{tabular}
\end{table*}

MSTAR-Public T72 Variants contain seven variants of the T72 in Table \ref{mstar2}.

\begin{table}[!htb]
\caption{Number of MSTAR-Public T72 Variants}
\label{mstar2}
\centering
\renewcommand\arraystretch{1.3}
\begin{tabular}{|c|c|c|c|c|c|c|c|c|c|} 
\hline
\multirow{3}{*}{DEG} & \multicolumn{9}{c|}{COL2} \\ 
\cline{2-10}
 & \multicolumn{8}{c|}{SCENE1} & SCENE3 \\ 
\cline{2-10}
 & A04 & A05 & A07 & A10 & A32 & A62 & A63 & A64 & A64 \\ 
\hline
15° & 274 & 274 & 274 & 271 & 274 & 274 & 274 & 274 &  \\ 
\hline
17° & 299 & 299 & 299 & 296 & 298 & 299 & 299 & 299 &  \\ 
\hline
30° &  &  &  &  &  &  &  & 288 & 133 \\ 
\hline
45° &  &  &  &  &  &  &  & 303 & 120 \\
\hline
\end{tabular}
\end{table}

MSTAR-Public Target Chips contain three public target variants (T72, BMP2, BTR70) and reference target (SLICY) in Table \ref{mstar3}.

\begin{table}[!htb]
\caption{Number of MSTAR-Public Target Chips}
\label{mstar3}
\centering
\renewcommand\arraystretch{1.3}
\begin{tabular}{|c|c|c|c|c|c|c|c|c|} 
\hline
\multirow{3}{*}{DEG} & \multicolumn{8}{c|}{hb} \\ 
\cline{2-9}
 & \multicolumn{3}{c|}{BMP2~ ~} & BTR70 & \multicolumn{3}{c|}{T72} & SLICY \\ 
\cline{2-9}
 & 9563 & 9566 & C21 & C71 & 132 & 812 & S7 &  \\ 
\hline
15° & 195 & 196 & 196 & 196 & 196 & 195 & 191 & 274 \\ 
\hline
17° & 233 & 232 & 233 & 233 & 232 & 231 & 228 &  \\ 
\hline
30° &  &  &  &  &  &  &  & 288 \\
\hline
\end{tabular}
\end{table}

\newpage
\section{Sample partitioning table} 
Based on the MSTAR dataset, researchers can establish different SOC and EOC settings. The following are detailed tables of our experimental setup: SOC in Table \ref{SOC}, EOC-Depression in Table \ref{Depression}, EOC-Configuration in Table \ref{Configuration}, EOC-Version in Table \ref{Version}, and EOC-Scene in Table \ref{Scene}. EOC-Scene/Gaussian/Random/Occlusion settings are based on the partitioning of the SOC.

\begin{table}[!htb]
\centering
\caption{Information of SOC}
\label{SOC}
\renewcommand\arraystretch{1.3}
\begin{tabular}{|c|c|c|c|c|} \hline
\multirow{2}{*}{Class} & Training & Test & \multirow{2}{*}{Collection} & \multirow{2}{*}{Scene} \\ \cline{2-3}
 & DEG/17° & DEG/15° &  &  \\ \hline
\begin{tabular}[c]{@{}c@{}}BMP2~\\(9563,~9566,~C21)\end{tabular} & 698 & 587 & \multicolumn{2}{c|}{} \\ \cline{1-3}
\begin{tabular}[c]{@{}c@{}}BTR70~\\(C71)\end{tabular} & 233 & 196 & \multicolumn{2}{c|}{hb} \\ \cline{1-3}
\begin{tabular}[c]{@{}c@{}}T72 \\(132,~812,~S7)\end{tabular} & 691 & 582 & \multicolumn{2}{c|}{} \\ \hline
BTR60 & 256 & 195 & COL1 & \multirow{7}{*}{SCENE1} \\ \cline{1-4}
2S1 & 299 & 274 & \multirow{6}{*}{COL2} &  \\ \cline{1-3}
BRDM2 & 298 & 274 &  &  \\ \cline{1-3}
D7 & 299 & 274 &  &  \\ \cline{1-3}
T-62 & 299 & 273 &  &  \\ \cline{1-3}
ZIL-131 & 299 & 274 &  &  \\ \cline{1-3}
ZSU234 & 299 & 274 &  &  \\ \hline
\end{tabular}
\end{table}

\begin{table}[!htb]
\centering
\caption{Information of EOC-Depression}
\label{Depression}
\renewcommand\arraystretch{1.3}
\begin{tabular}{|c|c|c|c|c|} 
\hline
\multirow{2}{*}{Class} & Training & Test & \multirow{2}{*}{Collection} & \multirow{2}{*}{Scene} \\ 
\cline{2-3}
 & DEG/17° & DEG/30° &  &  \\ 
\hline
2S1 & 299 & 288 & \multirow{4}{*}{COL2} & \multirow{4}{*}{SCENE1} \\ 
\cline{1-3}
BRDM2 & 298 & 287 &  &  \\ 
\cline{1-3}
ZSU234 & 299 & 288 &  &  \\ 
\cline{1-3}
T72 (A64) & 299 & 288 &  &  \\
\hline
\end{tabular}
\end{table}

\begin{table}[!htb]
\centering
\caption{Information of EOC-Configuration}
\label{Configuration}
\renewcommand\arraystretch{1.3}
\begin{tabular}{|c|c|c|c|c|c|c|c|c|} 
\hline
\multicolumn{4}{|c|}{Training} & \multicolumn{5}{c|}{Test} \\ 
\hline
Class & DEG/17° & Collection & Scene & \multicolumn{2}{c|}{Class} & DEG/15°,17° & Collection & Scene \\ 
\hline
BRDM2 & 298 & COL2 & SCENE1 & \multirow{5}{*}{T72} & A32 & 572 & \multirow{4}{*}{COL2} & \multirow{4}{*}{SCENE1} \\ 
\cline{1-4}\cline{6-7}
BMP2 (9563) & 233 & \multicolumn{2}{c|}{\multirow{3}{*}{hb}} &  & A62 & 573 &  &  \\ 
\cline{1-2}\cline{6-7}
BTR70 & 233 & \multicolumn{2}{c|}{} &  & A63 & 573 &  &  \\ 
\cline{1-2}\cline{6-7}
T72 (132) & 232 & \multicolumn{2}{c|}{} &  & A64 & 573 &  &  \\ 
\cline{1-4}\cline{6-9}
\multicolumn{4}{|c|}{} &  & S7 & 419 & \multicolumn{2}{c|}{hb} \\
\hline
\end{tabular}
\end{table}

\begin{table}[!htb]
\centering
\caption{Information of EOC-Version}
\label{Version}
\renewcommand\arraystretch{1.3}
\begin{tabular}{|c|c|c|c|c|c|c|c|c|} 
\hline
\multicolumn{4}{|c|}{Training} & \multicolumn{5}{c|}{Test} \\ 
\hline
Class & DEG/17° & Collection & Scene & \multicolumn{2}{c|}{Class} & DEG/15°,17° & Collection & Scene \\ 
\hline
BRDM2 & 298 & COL2 & SCENE1 & \multirow{2}{*}{BMP2} & 9566 & 428 & \multicolumn{2}{c|}{\multirow{2}{*}{hb}} \\ 
\cline{1-4}\cline{6-7}
BMP2 (9563) & 233 & \multicolumn{2}{c|}{\multirow{3}{*}{hb}} &  & C21 & 429 & \multicolumn{2}{c|}{} \\ 
\cline{1-2}\cline{5-9}
BTR70 & 233 & \multicolumn{2}{c|}{} & \multirow{5}{*}{T72} & A04 & 573 & \multirow{4}{*}{COL} & \multirow{4}{*}{SCENE1} \\ 
\cline{1-2}\cline{6-7}
T72 (132) & 232 & \multicolumn{2}{c|}{} &  & A05 & 573 &  &  \\ 
\cline{1-4}\cline{6-7}
\multicolumn{4}{|c|}{\multirow{3}{*}{}} &  & A07 & 573 &  &  \\ 
\cline{6-7}
\multicolumn{4}{|c|}{} &  & A10 & 567 &  &  \\ 
\cline{6-9}
\multicolumn{4}{|c|}{} &  & 812 & 426 & \multicolumn{2}{c|}{hb} \\
\hline
\end{tabular}
\end{table}

\begin{table}[!htb]
\centering
\caption{Information of EOC-Scene}
\label{Scene}
\renewcommand\arraystretch{1.3}
\begin{tabular}{|c|c|c|c|c|c|} 
\hline
\multirow{3}{*}{Class} & \multicolumn{5}{c|}{Collection/COL2} \\ 
\cline{2-6}
 & \multirow{2}{*}{DEG} & \multicolumn{2}{c|}{Training} & \multicolumn{2}{c|}{Testing} \\ 
\cline{3-6}
 &  & Scene & Number & Scene & Number \\ 
\hline
\multirow{2}{*}{BRDM-2} & 30° & \multirow{6}{*}{SCENE1} & 287 & \multirow{2}{*}{SCENE3} & 133 \\ 
\cline{2-2}\cline{4-4}\cline{6-6}
 & 45° &  & 303 &  & 120 \\ 
\cline{1-2}\cline{4-6}
\multirow{2}{*}{ZSU-23-4} & 30° &  & 288 & \multirow{2}{*}{SCENE2} & 118 \\ 
\cline{2-2}\cline{4-4}\cline{6-6}
 & 45° &  & 303 &  & 119 \\ 
\cline{1-2}\cline{4-6}
\multirow{2}{*}{T72(A64)} & 30° &  & 288 & \multirow{2}{*}{SCENE3} & 133 \\ 
\cline{2-2}\cline{4-4}\cline{6-6}
 & 45° &  & 303 &  & 120 \\
\hline
\end{tabular}
\end{table}

\bibliographystyle{IEEEtran}
\bibliography{ref}